%% file: main.tex
\documentclass[10pt,twocolumn,letterpaper]{article}

\usepackage[pagenumbers]{cvpr} 

\input{preamble}

\definecolor{cvprblue}{rgb}{0.21,0.49,0.74}
\usepackage[pagebackref,breaklinks,colorlinks,citecolor=cvprblue]{hyperref}

\def\paperID{11709} 
\def\confName{CVPR}
\def\confYear{2024}

\def\method{{DiverGen}}

\title{\method: Improving Instance Segmentation by Learning Wider Data Distribution with More Diverse Generative Data}

\author{
Chengxiang Fan$^1$\thanks{Equal contribution.}~~
Muzhi Zhu$^1$\footnotemark[1]~~
Hao Chen$^1$\thanks{Correspondence should be addressed to HC and CS.}~~
Yang Liu$^1$~~
Weijia Wu$^1$~~
Huaqi Zhang$^2$~~
Chunhua Shen$^1$\footnotemark[2]
\\[.25cm]
$ ^1$ Zhejiang University, China 
~~~~~
$ ^2$ 
vivo Mobile Communication Co.
}

\begin{document}
\maketitle
\input{sec/0_abstract}    
\input{sec/1_intro}

\input{sec/2_related}
\input{sec/3_method}
\input{sec/4_experiment}
\input{sec/5_conclusion}

{
    \small
    \bibliographystyle{ieeenat_fullname}
    \bibliography{main}
}

\input{sec/X_suppl}

\end{document}

%% file: preamble.tex
\usepackage[dvipsnames]{xcolor}

\usepackage{xcolor}
\usepackage{colortbl}
\usepackage{bm}
\usepackage{multicol}
\usepackage{multirow}
\usepackage{makecell}
\usepackage{longtable}
\usepackage{microtype}
\usepackage[accsupp]{axessibility} 

%% file: sec/0_abstract.tex
\begin{abstract}
Instance segmentation is data-hungry, and as model capacity increases, data scale becomes crucial for 
improving the accuracy.
Most instance segmentation datasets today require costly manual annotation, limiting their data scale. Models trained on such data are prone to overfitting on the training set, especially for those rare categories. While recent works have delved into 
exploiting 
generative models to create synthetic datasets for data augmentation, these approaches do not efficiently harness the full potential of generative models. 

To address these issues, 
we 
introduce a more efficient strategy to construct generative datasets for data augmentation, termed \textbf{\method}. Firstly, we provide an explanation of the role of generative data from the perspective of distribution discrepancy. We investigate the impact of different data on the distribution learned by the model. We argue that generative data can expand the data distribution that the model can learn,
thus mitigating overfitting. 
Additionally, we find that the diversity of generative data is crucial for improving model performance and enhance it through various strategies, including category diversity, prompt diversity, and generative model diversity. 
With 
these strategies, we can scale the data to millions while maintaining the trend of model performance improvement. On the LVIS dataset, \method\ significantly outperforms the strong model X-Paste, achieving {+$\bf 1.1$} box AP and {+$\bf 1.1$} mask AP across all categories, and {+$\bf 1.9$} box AP and {+$\bf 2.5$} mask AP for rare categories.
Our codes are available at \href{https://github.com/aim-uofa/DiverGen}{https://github.com/aim-uofa/DiverGen}.
\end{abstract}

%% file: sec/1_intro.tex
\section{Introduction}
\label{sec:intro}

Instance segmentation \cite{he2017mask, cheng2022masked, carion2020end} is one of the challenging tasks in computer vision, requiring the prediction of masks and categories for instances in an image, which serves as the foundation for numerous visual applications. As models' learning capabilities improve, the demand for training data increases. However, current datasets for instance segmentation heavily rely on manual annotation, which is time-consuming and costly, and the dataset scale cannot meet the training needs of models. 
Despite the recent emergence of the automatically annotated dataset SA-1B~\cite{kirillov2023segment}, it lacks category annotations, failing to meet the requirements of instance segmentation.
Meanwhile, the ongoing development of the generative model has 
largely improved 
the controllability and realism of 
generated samples. 
For example, the recent 
text2image diffusion model~\cite{rombach2021highresolution, alex2023deepfloyd} can generate high-quality images corresponding to input prompts. Therefore, current methods~\cite{zhao2022xpaste, wu2023diffumask, wu2023datasetdm} use generative models for data augmentation by generating datasets to supplement the training of models on real datasets and improve model performance. Although current methods have proposed various strategies to enable generative data to boost model performance, there are still some limitations: 
1) Existing methods have not fully exploited the potential of generative models. First, some methods~\cite{zhao2022xpaste} not only use generative data but also need to crawl images from the internet, which is significantly challenging to obtain large-scale data. Meanwhile, the content of data crawled from the internet is uncontrollable and needs extra checking. Second, existing methods do not fully use the controllability of generative models. Current methods often adopt manually designed templates to construct prompts, limiting the potential output of generative models. 
2) Existing methods~\cite{wu2023diffumask, wu2023datasetdm} often explain the role of generative data from the perspective of class imbalance or data scarcity, without considering the discrepancy between real-world data and generative data. Moreover, these methods typically show improved model performance only in scenarios with a limited number of real samples, and the effectiveness of generative data on existing large-scale real datasets, like LVIS~\cite{gupta2019lvis}, is not thoroughly investigated.

In this paper, we  
first 
explore the role of generative data from the perspective of distribution discrepancy, addressing two main questions:
1) \textit{Why does generative data augmentation enhance model performance?}
2) \textit{What types of generative data are beneficial for improving model performance?
} 
First, we find that there exist discrepancies between the model learned distribution of the limited real training data and the distribution of real-world data. We visualize the data and find that compared to the real-world data, generative data can expand the data distribution that the model can learn. Furthermore, we find that the role of adding generative data is to alleviate the bias of the real training data, effectively mitigating overfitting the training data. 
Second, we find that there are also discrepancies between the distribution of the generative data and the real-world data distribution. If these discrepancies are not handled properly, the full potential of the generative model cannot be utilized. By conducting several experiments, we find that using diverse generative data enables models to better adapt to these discrepancies, improving model performance.

Based on the above analysis, we propose an efficient strategy for enhancing data diversity, namely, \textit{Generative Data Diversity Enhancement}. We design various diversity enhancement strategies to increase data diversity from the perspectives of \textit{category diversity, prompt diversity, and generative model diversity}. 
For category diversity, we observe that models trained with generative data covering all categories adapt better to distribution discrepancy than models trained with partial categories. Therefore, we introduce not only categories from LVIS~\cite{gupta2019lvis} but also extra categories from ImageNet-1K~\cite{russakovsky2015imagenet} to enhance category diversity in data generation, thereby reinforcing the model's adaptability to distribution discrepancy. 
For prompt diversity, we find that as the scale of the generative dataset increases, manually designed prompts cannot scale up to the corresponding level, limiting the diversity of output images from the generative model. Thus, we design a set of diverse prompt generation strategies to use large language models, like ChatGPT, for prompt generation, requiring the large language models to output maximally diverse prompts under constraints. By combining manually designed prompts and ChatGPT designed prompts, we effectively enrich prompt diversity and further improve generative data diversity.
For generative model diversity, we find that data from different generative models also exhibit distribution discrepancies. Exposing models to data from different generative models during training can enhance adaptability to different distributions. Therefore, we employ Stable Diffusion~\cite{rombach2021highresolution} and DeepFloyd-IF~\cite{alex2023deepfloyd} to generate images for all categories separately and mix the two types of data during training to increase data diversity.

At the same time, we optimize the data generation workflow and propose a four-stage generative pipeline consisting of instance generation, instance annotation, instance filtration, and instance augmentation.
In the instance generation stage, we employ our proposed Generative Data Diversity Enhancement to enhance data diversity, producing diverse raw data. 
In the instance annotation stage, we introduce an annotation strategy called SAM-background. This strategy obtains high-quality annotations by using background points as input prompts for SAM~\cite{kirillov2023segment}, obtaining the annotations of raw data.
In the instance filtration stage, we introduce a metric called CLIP inter-similarity. Utilizing the CLIP~\cite{radford2021learning} image encoder, we extract embeddings from generative and real data, and then compute their similarity. A lower similarity indicates lower data quality. After filtration, we obtain the final generative dataset.
In the instance augmentation stage, we use the instance paste strategy~\cite{zhao2022xpaste} to increase model learning efficiency on generative data.

Experiments demonstrate that our designed data diversity strategies can effectively improve model performance and maintain the trend of performance gains as the data scale increases to the million level, which enables large-scale generative data for data augmentation. On the LVIS dataset, \method\ significantly outperforms the strong model X-Paste~\cite{zhao2022xpaste}, achieving {+$\bf 1.1$} box AP~\cite{gupta2019lvis} and {+$\bf 1.1$} mask AP across all categories, and {+$\bf 1.9$} box AP and {+$\bf 2.5$} mask AP for rare categories.

In summary, 
our main 
contributions are as follows: 
\begin{itemize}
\item We explain the role of generative data from the perspective of distribution discrepancy. We find that generative data can expand the data distribution that the model can learn, mitigating overfitting the training set and the diversity of generative data is crucial for improving model performance. 
\item We propose the Generative Data Diversity Enhancement strategy to increase data diversity from the aspects of category diversity, prompt diversity, and generative model diversity. By enhancing data diversity, we can scale the data to millions
while maintaining the trend of model performance improvement.
\item We optimize the data generation pipeline. We propose an annotation strategy SAM-background to obtain higher-quality annotations. We also introduce a filtration metric called CLIP inter-similarity to filter data and further improve the quality of the generative dataset. 
\end{itemize}

%% file: sec/2_related.tex
\section{Related Work}
\label{sec:related}
\noindent\textbf{Instance segmentation.}
Instance segmentation is an important task in the field of computer vision and has been extensively studied. Unlike semantic segmentation, instance segmentation not only classifies the pixels at a pixel level but also distinguishes different instances of the same category. Previously, the focus of instance segmentation research has primarily been on the design of model structures. Mask-RCNN~\cite{he2017mask} unifies the tasks of object detection and instance segmentation. Subsequently, Mask2Former~\cite{cheng2022masked} further unified the tasks of semantic segmentation and instance segmentation by leveraging the structure of DETR~\cite{carion2020end}.

\begin{figure*}[t]
  \centering
  \includegraphics[width=0.8\linewidth]{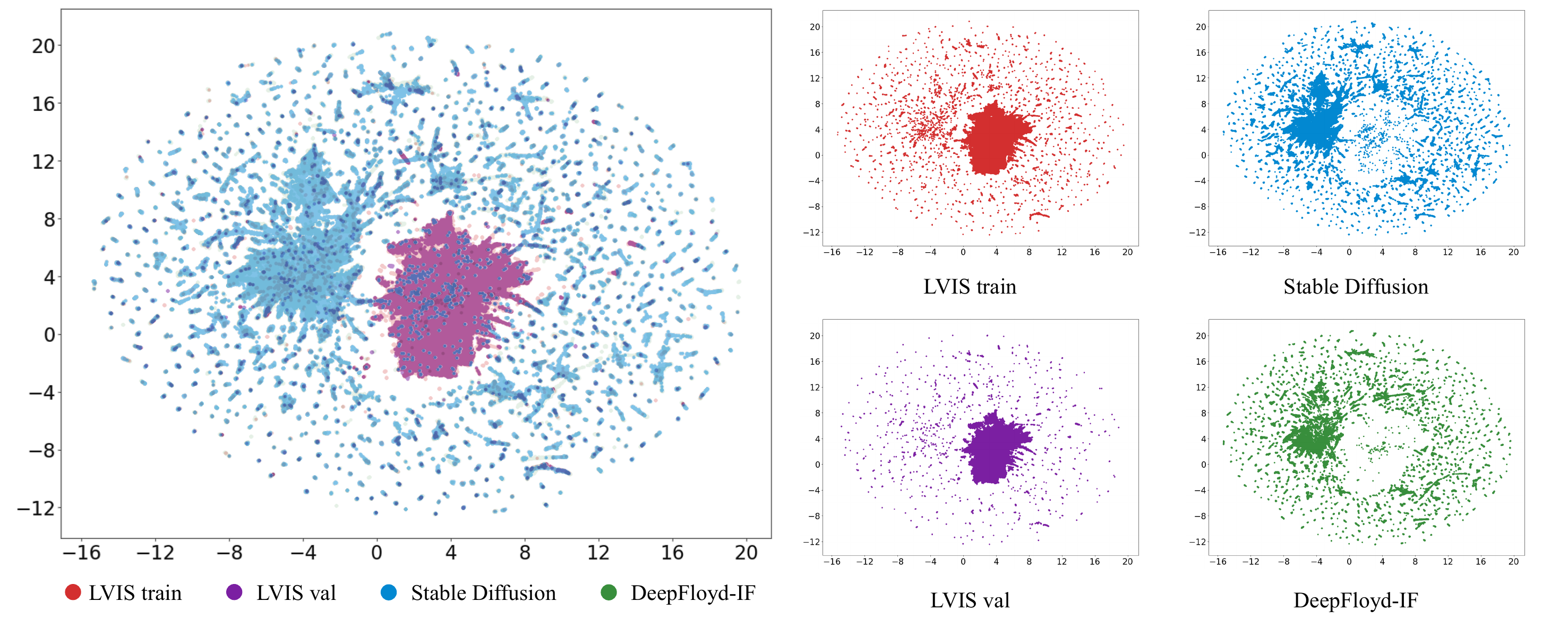}
  \caption{\textbf{Visualization of data distributions on different sources.}
  Compared to real-world data (LVIS train and LVIS val), generative data (Stable Diffusion and IF) can expand the data distribution that the model can learn.}
  \label{fig:dist}
\end{figure*}

Orthogonal to these studies focusing on model architecture, our work primarily investigates how to better utilize generated data for this task.  We focus on the challenging long-tail dataset LVIS \cite{gupta2019lvis} because it is only the long-tailed categories that face the issue of limited real data and require generative images for augmentation, making it more practically meaningful.

\noindent\textbf{Generative data augmentation.}
 The use of generative models to synthesize training data for assisting perception tasks such as classification~\cite{feng2023diverse,zhang2023prompt}, detection~\cite{zhao2022xpaste,chen2023integrating}, segmentation~\cite{li2023open,wu2023diffumask,wu2023datasetdm}, etc. has received widespread attention from researchers.  
 In the field of segmentation, early works~\cite{zhang2021datasetgan,li2022bigdatasetgan}  utilize generative adversarial networks (GANs) to synthesize additional training samples. With the rise of diffusion models,  there have been numerous efforts~\cite{zhao2022xpaste,li2023open,wu2023diffumask,wu2023datasetdm,yang2023freemask}  to utilize text2image diffusion models, such as Stable Diffusion~\cite{rombach2021highresolution}, to boost the segmentation performance. \citet{li2023open} combine the Stable Diffusion model with a novel grounding module and establish an automatic pipeline for constructing a segmentation dataset. DiffuMask 
\cite{wu2023diffumask} exploits the potential of cross-attention maps between text and images to synthesize accurate semantic labels. More recently, FreeMask~\cite{yang2023freemask} uses a mask-to-image generation model to generate images conditioned on the provided semantic masks. However, the aforementioned work is only applicable to semantic segmentation. The most relevant work to ours is X-Paste \cite{zhao2022xpaste}, which promotes instance segmentation through copy-pasting the generative images and a filter strategy based on CLIP~\cite{radford2021learning}.

In summary, most methods only demonstrate significant advantages when training data is extremely limited. They consider generating data as a means to compensate for data scarcity or class imbalance. However, in this work, we take a further step to examine and analyze this problem from the perspective of data distribution. We propose a pipeline that enhances diversity from multiple levels to alleviate the impact of data distribution discrepancies. 
This provides new insights and inspirations for further advancements in this field.

%% file: sec/3_method.tex
\section{Our Proposed \method}

\subsection{Analysis of Data Distribution}
Existing methods~\cite{zhao2022xpaste,wu2023diffumask,xie2023mosaicfusion} often attribute the role of generative data to addressing class imbalance or data scarcity. In this paper, we provide an explanation for two main questions from the perspective of distribution discrepancy.

\noindent \textbf{Why does generative data augmentation enhance model performance?}
We argue that there exist discrepancies between the model learned distribution of the limited real training data and the distribution of real-world data. The role of adding generative data is to alleviate the bias of the real training data, effectively mitigating overfitting the training data.

First, to intuitively understand the discrepancies between different data sources, we use CLIP~\cite{radford2021learning} image encoder to extract the embeddings of images from different data sources, and then use UMAP~\cite{mcinnes2018umap} to reduce dimensions for visualization. Visualization of data distributions on different sources is shown in Figure~\ref{fig:dist}. 
Real-world data (LVIS~\cite{gupta2019lvis} train and LVIS val) cluster near the center, while generative data (Stable Diffusion~\cite{rombach2021highresolution} and IF~\cite{alex2023deepfloyd}) are more dispersed, indicating that generative data can expand the data distribution that the model can learn.

Then, to characterize the distribution learned by the model, we employ the free energy formulation used by \citet{joseph2021towards}. This formulation transforms the logits outputted by the classification head into an energy function. The formulation is shown below:
\begin{equation}
    F(\bm{q}; h) = -\tau \log \sum_{c = 1}^{n} \exp
    \left(
    \frac{h_c(\bm{q})}{\tau}
    \right).
    \label{eqn:function}
\end{equation}
\noindent Here, $\bm{q}$ is the feature of instance, $h_c(\bm{q})$ is the $c^{th}$ logit outputted by classification head $h(.)$, $n$ is the number of categories and $\tau$ is the temperature parameter. 
We train one model using only the LVIS train set ($\theta_{\text{train}}$), and another model using LVIS train with generative data ($\theta_{\text{gen}}$). Both models are evaluated on the LVIS val set and we use instances that are successfully matched by both models to obtain energy values. Additionally, we train another model using LVIS val ($\theta_{\text{val}}$), treating it as representative of real-world data distribution.  
Then, we further fit Gaussian distributions to the histograms of energy values to obtain the mean $\mu$ and standard deviation $\sigma$ for each model and compute the KL divergence~\cite{joyce2011kullback} between them. $D_{KL}(p_{\theta_\text{train}}\|p_{\theta_\text{val}})$ is 0.063, and $D_{KL}(p_{\theta_\text{gen}}\|p_{\theta_\text{val}})$ is 0.019. The latter is lower, indicating that using generative data mitigates the bias of limited real training data. 

Moreover, we also analyze the role of generative data from a metric perspective. We randomly select up to five images per category to form a minitrain set and then conduct inferences using $\theta_{\text{train}}$ and $\theta_{\text{gen}}$. Then, we define a metric, 
termed 
train-val gap (\text{TVG}), which is formulated as follows:
\begin{equation}
    \text{TVG}_w^k = \text{AP}_{w}^{k}minitrain - \text{AP}_{w}^{k}val.
    \label{eqn:tvg}
\end{equation}

\noindent Here, $\text{TVG}_w^k$ is train-val gap of $w$ category on task $k$, $\text{AP}_{w}^{k}d$ is AP~\cite{gupta2019lvis} of $w$ category on $k$ obtained on dataset $d$, $w \in \{f, c, r\}$, with $f$, $c$, $r$ standing for frequent, common, rare~\cite{gupta2019lvis} respectively, and $k\in \{box, mask\}$, with $box$, $mask$ referring to the object detection and instance segmentation.
The train-val gap serves as a measure of the disparity in the model's performance between the training and validation sets. A larger gap indicates a higher degree of overfitting the training set.
The results, as presented in Table~\ref{tab:source}, show that the metrics for the rare categories consistently surpass those of frequent and common. This observation suggests that the model tends to overfit more on the rare categories that have fewer examples. With the augmentation of generative data, all \text{TVG} of $\theta_{\text{gen}}$ are lower than $\theta_{\text{train}}$, showing that adding generative data can effectively alleviate overfitting the training data.

\begin{table}[h]
  \small
  \centering
  \resizebox{\linewidth}{!}{
      \begin{tabular}{c|cccccccc}
      \toprule
      Data Source&$\text{TVG}_f^{box}$&$\text{TVG}_f^{mask}$&$\text{TVG}_c^{box}$&$\text{TVG}_c^{mask}$&$\text{TVG}_r^{box}$&$\text{TVG}_r^{mask}$ \\
      \midrule
      LVIS  & 13.16  & 10.71  & 21.80  & 16.80  & 39.59  & 31.68 \\
      LVIS + Gen & 9.64  & 8.38  & 15.64  & 12.69  & 29.39  & 22.49 \\
    \bottomrule
    \end{tabular}
    }

\caption{\textbf{Results of train-val gap on different data sources.}
With the augmentation of generative data, all \text{TVG} of LVIS are lower than LVIS + Gen, showing that adding generative data can effectively alleviate overfitting to the training data.}
\label{tab:source}
\end{table}

\noindent \textbf{What types of generative data are beneficial for improving model performance?}
We argue 
that  
there are also discrepancies between the distribution of the generative data and the real-world data distribution. If these discrepancies are not 
properly addressed, the full potential of the generative model cannot be  
attained.

We divide the generative data into `frequent', `common', and `rare'~\cite{gupta2019lvis} groups, and train three models using each group of data as instance paste source. The inference results are shown in Table~\ref{tab:subset}. We find that the metrics on the corresponding category subset are lowest when training with only one group of data.  
We consider model performance to be primarily influenced by the quality and diversity of data. Given that the quality of generative data is relatively consistent, we contend insufficient diversity in the data can mislead the distribution that the model can learn and a more comprehensive understanding is obtained by the model from a diverse set of data. Therefore, we believe 
that 
\textit{using diverse generative data enables models to better adapt to these discrepancies}, improving model performance.

\begin{table}[h]
  \small
  \centering
  \resizebox{\linewidth}{!}{
      \begin{tabular}{c|ccccccccc}
      \toprule
       \# Gen Category&$\text{AP}_f^{box}$&$\text{AP}_f^{mask}$&$\text{AP}_c^{box}$&$\text{AP}_c^{mask}$&$\text{AP}_r^{box}$&$\text{AP}_r^{mask}$ \\
      \midrule
      none & 50.14  & 43.84  & 47.54  & 43.12  & 41.39  & 36.83  \\ 
      f & \textcolor{blue}{50.81}  & \textcolor{blue}{44.24}  & 47.96  & 43.51  & 41.51  & 37.92  \\ 
      c & 51.86  & 45.22  & \textcolor{blue}{47.69}  & \textcolor{blue}{42.79}  & 42.32  & 37.30  \\ 
      r & 51.46  & 44.90  & 48.24  & 43.51  & \textcolor{blue}{32.67}  & \textcolor{blue}{29.04}  \\ 
      all & 52.10  & 45.45  & 50.29  & 44.87  & 46.03  & 41.86 \\
    \bottomrule
    \end{tabular}
    }
\vspace{-2mm}
\caption{\textbf{Results of different category data subset for training.}
The metrics on the corresponding category subset are lowest when training with only one group of data,  
showing insufficient diversity in the data can mislead the distribution that the model can learn.
\textcolor{blue}{Blue} font means the lowest value in models using generative data.}
\label{tab:subset}
\end{table}

\subsection{Generative Data Diversity Enhancement}
\label{subsec:gdde}
Through the analysis above, we find that the diversity of generative data is crucial for improving model performance. Therefore, we design a series of strategies to enhance data diversity at three levels: category diversity, prompt diversity, and generative model diversity, which help the model to better adapt to the distribution discrepancy between generative data and real data.

\noindent\textbf{Category diversity.}
The above experiments show that including data from partial categories results in lower performance than incorporating data from all categories. We believe that, akin to human learning, the model can learn features beneficial to the current category from some other categories. Therefore, we consider increasing the diversity of data by adding extra categories.
First, we select some extra categories besides LVIS from ImageNet-1K~\cite{russakovsky2015imagenet} categories based on WordNet~\cite{fellbaum2010wordnet} similarity. Then, the generative data from LVIS and extra categories are mixed for training, requiring the model to learn to distinguish all categories. Finally, we truncate the parameters in the classification head corresponding to the extra categories during inference, ensuring that the inferred category range remains within LVIS.

\noindent\textbf{Prompt diversity.}
The output images of the text2image generative model typically rely on the input prompts. Existing methods~\cite{zhao2022xpaste} usually generate prompts by manually designing templates, such as ``a photo of a single \textit{\{category\_name\}}." When the data scale is small, designing prompts manually is convenient and fast. However, when generating a large scale of data, it is challenging to scale the number of manually designed prompts correspondingly. 
Intuitively, it is essential to diversify the prompts to enhance data diversity. To easily generate a large number of prompts, we choose large language model, like ChatGPT, to enhance the prompt diversity. We have three requirements for the large language model: 1) each prompt should be as different as possible; 2) each prompt should ensure that there is only one object in the image; 3) prompts should describe different attributes of the category. For example, if the category is food, prompts should cover attributes like color, brand, size, freshness, packaging type, packaging color, etc. Limited by the inference cost of ChatGPT, we use the manually designed prompts as the base and only use ChatGPT to enhance the prompt diversity for a subset of categories. 
Moreover, we also leverage the controllability of the generative model, adding the constraint ``in a white background" after each prompt to make the background of output images simple and clear, which reduces the difficulty of mask annotation.

\begin{figure}[h]
  \centering
  \includegraphics[width=\linewidth]{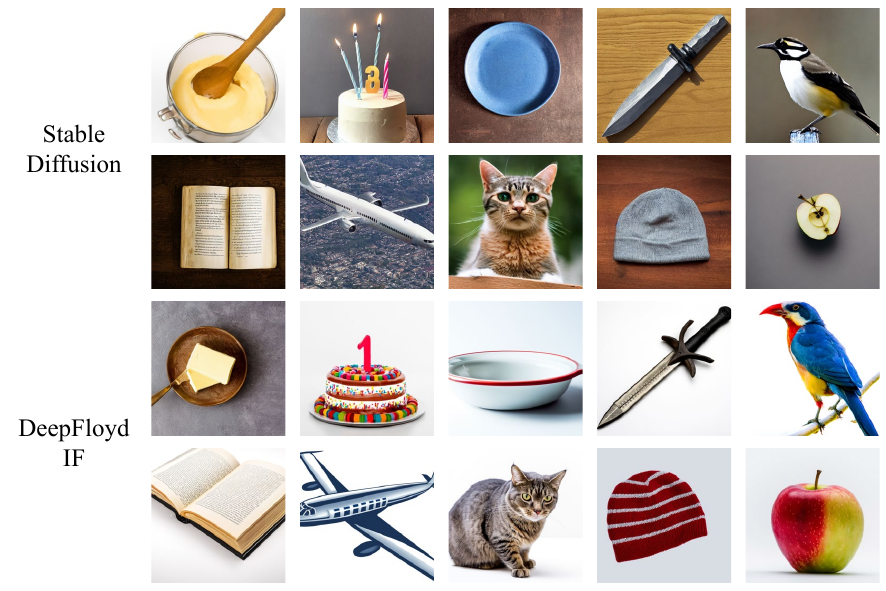}
  \caption{\textbf{Examples 
  of various 
  generative models.}
  The samples generated by different generative models vary, even within the same category.}
  \label{fig:model}
\end{figure}

\noindent\textbf{Generative model diversity.}
The quality and style of output images vary across generative models, and the data distribution learned solely from one generative model's data is limited. Therefore, we introduce multiple generative models to enhance the diversity of data, allowing the model to learn from wider data distributions. We selected two commonly used generative models, Stable Diffusion~\cite{rombach2021highresolution} (SD) and DeepFloyd-IF~\cite{alex2023deepfloyd} (IF). We use Stable Diffusion  V1.5, generating images with a resolution of 512~$\times$~512, and use images output from Stage II of IF with a resolution of 256~$\times$~256. For each category in LVIS, we generated 1k images using two models separately. Examples from different generative models are shown in Figure~\ref{fig:model}.

\label{sec:method}
\begin{figure*}
  \centering
  \includegraphics[width=\linewidth]{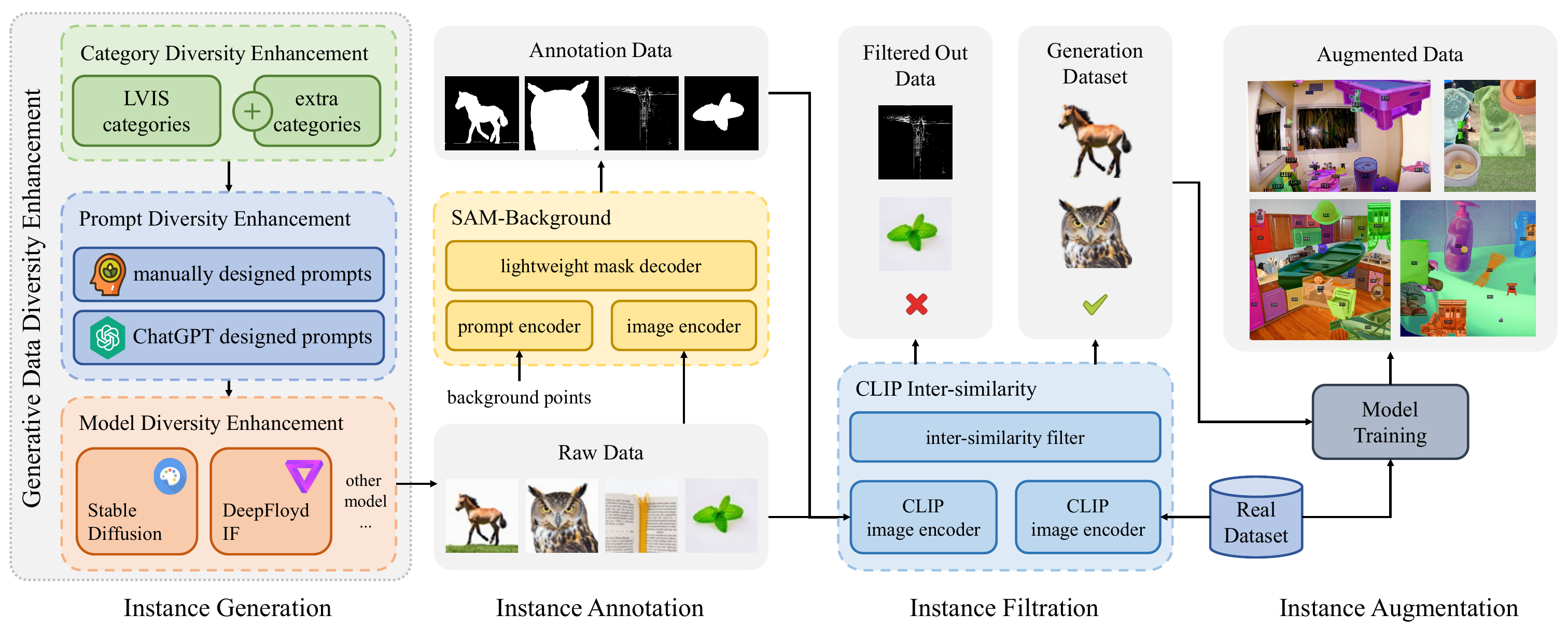}
  \caption{\textbf{Overview of the \method\ pipeline.} 
  In instance generation, we enhance data diversity at three levels: category diversity, prompt diversity, and generative model diversity. Next, we use SAM-background to obtain high-quality masks. Then, we use CLIP inter-similarity to filter out low-quality data. At last, we use the instance paste strategy to increase model learning efficiency on generative data.}
  \label{fig:overview}
\end{figure*}

\subsection{Generative Pipeline}
The generative pipeline of \method\ is built upon X-Paste~\cite{zhao2022xpaste}. It can be divided into four stages: instance generation, instance annotation, instance filtration and instance augmentation. The overview of \method\ is illustrated in Figure~\ref{fig:overview}.

\noindent\textbf{Instance generation.}
Instance generation is a crucial stage for enhancing data diversity. In this stage, we employ our proposed Generative Data Diversity Enhancement (GDDE), as mentioned in Sec~\ref{subsec:gdde}. In category diversity enhancement, we utilize the category information from LVIS~\cite{gupta2019lvis} categories and extra categories selected from ImageNet-1K~\cite{russakovsky2015imagenet}. In prompt diversity enhancement, we utilize manually designed prompts and ChatGPT designed prompts to enhance prompt diversity. In model diversity enhancement, we employ two generative models, SD and IF.

\noindent\textbf{Instance annotation.}
We employ SAM~\cite{kirillov2023segment} as our annotation model. SAM is a class-agnostic promptable segmenter that outputs corresponding masks based on input prompts, such as points, boxes, etc. In instance generation, leveraging the controllability of the generative model, the generative images have two characteristics: 1) each image predominantly contains only one foreground object; 2) the background of the images is relatively simple. Therefore, we introduce a SAM-background (SAM-bg) annotation strategy. SAM-bg takes the four corner points of an image as input prompts for SAM to obtain the background mask, then inverts the background mask as the mask of the foreground object. Due to the conditional constraints during the instance generation stage, this strategy is simple but effective in producing high-quality masks.

\noindent\textbf{Instance filtration.}
In the instance filtration stage, X-Paste utilizes the CLIP score (similarity between images and text) as the metric for image filtering. However, we observe that the CLIP score is ineffective in filtering low-quality images. In contrast to the similarity between images and text, we think the similarity between images can better filter out low-quality images. Therefore, we propose a new metric called CLIP inter-similarity. We use the image encoder of CLIP~\cite{radford2021learning} to extract image embeddings for objects in the training set and generative images, then calculate the similarity between them. If the similarity is too low, it indicates a significant disparity between the generative and real images, suggesting that it is  probably a poor-quality image and needs to be filtered.

\noindent\textbf{Instance augmentation.}
We use the augmentation strategy proposed by X-Paste~\cite{zhao2022xpaste} but do not use the data retrieved from the network or the instances in LVIS~\cite{gupta2019lvis} training set as the paste data source, only use the generative data as the paste data source.

%% file: sec/4_experiment.tex
\section{Experiments}
\label{sec:experiment}

\subsection{Settings}

\noindent\textbf{Datasets.}
We choose LVIS~\cite{gupta2019lvis} for our experiments. LVIS is a large-scale instance segmentation dataset, containing 164k images with approximately two million high-quality annotations of instance segmentation and object detection. LVIS dataset uses images from COCO 2017~\cite{lin2014microsoft} dataset, but redefines the train/val/test splits, with around 100k images in the training set and around 20k images in the validation set. The annotations in LVIS cover 1,203 categories, with a typical long-tailed distribution of categories, so LVIS further divides the categories into frequent, common, and rare based on the frequency of each category in the dataset. We use the official LVIS training split and the validation split.

\noindent\textbf{Evaluation metrics.}
The evaluation metrics are LVIS box average precision ($\text{AP}^{box}$) and mask average precision ($\text{AP}^{mask}$). We also provide the average precision of rare categories ($\text{AP}_r^{box}$ and $\text{AP}_r^{mask}$). The maximum number of detections per image is 300.

\noindent\textbf{Implementation details.}
We use CenterNet2~\cite{zhou2021probabilistic} as the baseline and Swin-L~\cite{liu2021Swin} as the backbone. In the training process, we initialize the parameters by the pre-trained Swin-L weights provided by \citet{liu2021Swin}. The training size is 896 and the batch size is 16. The maximum training iterations is 180,000 with an initial learning rate of 0.0001. We use the instance paste strategy provided by \citet{zhao2022xpaste}. 

\subsection{Main Results}

\begin{figure*}[h]
  \centering
  \begin{subfigure}{0.49\linewidth}
    \includegraphics[width=\linewidth]{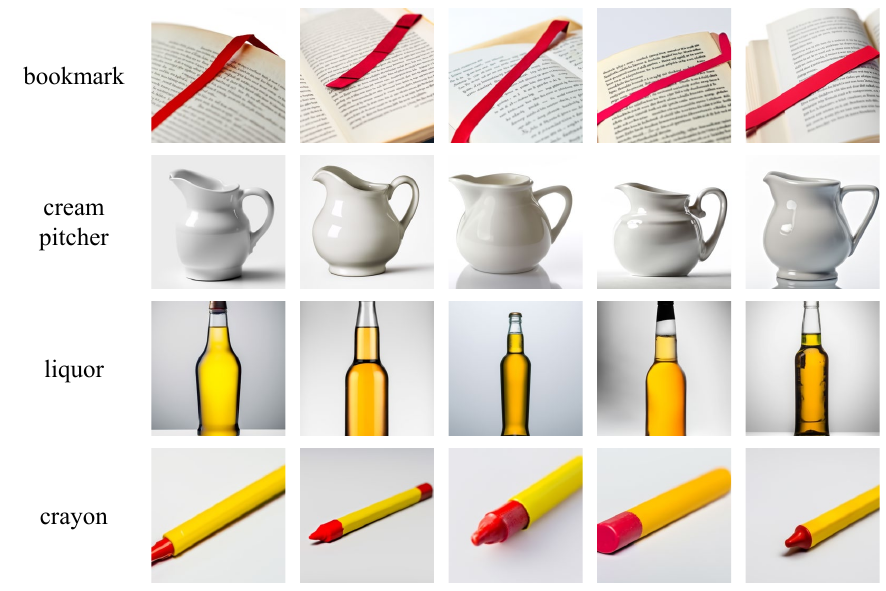}
    \caption{Images of manually designed prompts.}
    \label{fig:one_prompt}
  \end{subfigure}
  \hfill
  \begin{subfigure}{0.49\linewidth}
    \includegraphics[width=\linewidth]{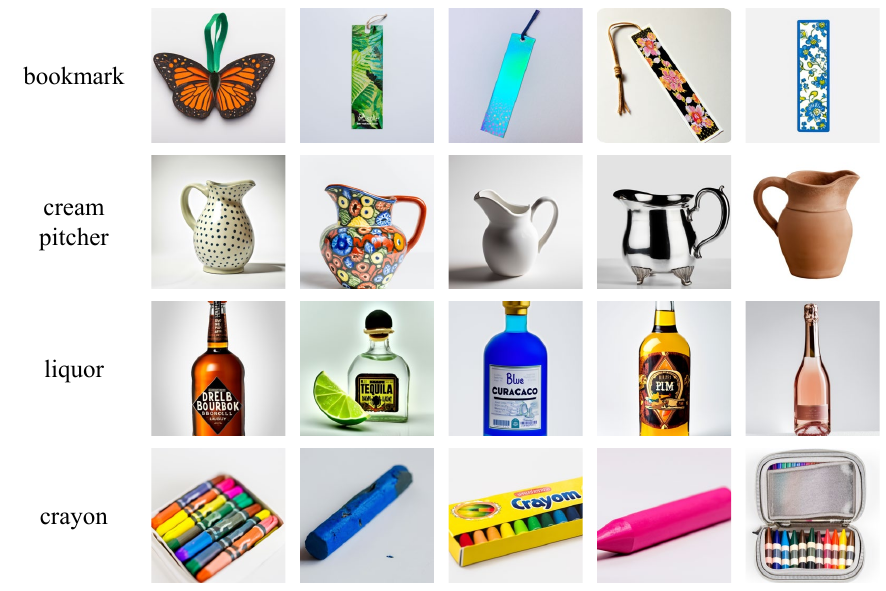}
    \caption{Images of ChatGPT designed prompts.}
    \label{fig:gpt_prompt}
  \end{subfigure}
  \caption{\textbf{Examples of generative data using different prompts.}
  By using prompts designed by ChatGPT, the diversity of generated images in terms of shapes, textures, etc. can be significantly improved. }
  \label{fig:prompt}
\end{figure*}

\noindent\textbf{Data diversity is more important than quantity.}
To investigate the impact of different scales of generative data, we use generative data of varying scales as paste data sources. We construct three datasets using only DeepFloyd-IF~\cite{alex2023deepfloyd} with manually designed prompts, all containing original LVIS 1,203 categories, but with per-category quantities of 0.25k, 0.5k, and 1k, resulting in total dataset scales of 300k, 600k, and 1,200k. As shown in Table~\ref{tab:gdd}, we find that using generative data improves model performance compared to the baseline. However, as the dataset scale increases, the model performance initially improves but then declines. The model performance using 1,200k data is lower than that using 600k data. Due to the limited number of manually designed prompts, the generative model produces similar data, as shown in Figure~\ref{fig:one_prompt}. Consequently, the model can not gain benefits from more data. However, when using our proposed Generative Data Diversity Enhancement (GDDE), due to the increased data diversity, the model trained with 1,200k images achieves better results than using 600k images, with an improvement of 1.21 box AP and 1.04 mask AP. Moreover, when using the same data scale of 600k, the mask AP increased by 0.64 AP and the box AP increased by 0.55 AP when using GDDE compared to not using it. The results demonstrate that data diversity is more important than quantity. When the scale of data is small, increasing the quantity of data can improve model performance, which we consider is an indirect way of increasing data diversity. However, this simplistic approach of solely increasing quantity to increase diversity has an upper limit. When it reaches this limit, explicit data diversity enhancement strategies become necessary to maintain the trend of model performance improvement.

\begin{table}[h]
  \small
  \centering
  \resizebox{0.9\linewidth}{!}{
      \begin{tabular}{c|c|cccc}
      \toprule
      \# Gen Data&GDDE&$\text{AP}^{box}$&$\text{AP}^{mask}$&$\text{AP}_r^{box}$&$\text{AP}_r^{mask}$ \\
      \midrule
      0 & ~ & 47.50  & 42.32  & 41.39  & 36.83 \\
      300k & ~ & 49.65  & 44.01  & 45.68  & 41.11 \\
      600k & ~ & 50.03  & 44.44  & 47.15  & 41.96 \\
      1200k & ~ & 49.44  & 43.75  & 42.96  & 37.91 \\
      600k & \checkmark & 50.67  & 44.99  & 48.52  & 43.63 \\
      1200k & \checkmark & \textbf{51.24}  & \textbf{45.48}  & \textbf{50.07}  & \textbf{45.85} \\
    \bottomrule
    \end{tabular}
    }
\vspace{-2mm}
\caption{\textbf{Results of different scales of generative data.} When using the same data scale, models using our proposed GDDE can achieve higher performance than those without it, showing that data diversity is more important than quantity.}
\label{tab:gdd}
\vspace{-3mm}
\end{table}

\noindent\textbf{Comparision with previous methods.}
We compare \method\ with previous data-augmentation related methods in Table~\ref{tab:main}. Compared to the baseline CenterNet2~\cite{zhou2021probabilistic}, our method significantly improves, increasing box AP by +3.7 and mask AP by +3.2. Regarding rare categories, our method surpasses the baseline with +8.7 in box AP and +9.0 in mask AP. Compared to the previous strong model X-Paste~\cite{zhao2022xpaste}, we outperform it with +1.1 in box AP and +1.1 in mask AP of all categories, and +1.9 in box AP and +2.5 in mask AP of rare categories. It is worth mentioning that, X-Paste utilizes both generative data and web-retrieved data as paste data sources during training, while our method exclusively uses generative data as the paste data source. We achieve this by designing diversity enhancement strategies, further unlocking the potential of generative models.

\begin{table}[h]
  \small
  \centering
  \resizebox{\linewidth}{!}{
    \begin{tabular}{r |c|cccc}
      \toprule
      Method & Backbone &AP$^{box}$&AP$^{mask}$&AP$_r^{box}$&AP$_r^{mask}$\\
      \midrule
      Copy-Paste \cite{ghiasi2021simple} &  EfficientNet-B7 &  41.6&38.1& -&32.1\\
      Tan et al.\ \cite{tan20201st} & ResNeSt-269  & - & 41.5 & - & 30.0 \\ 
      Detic \cite{zhou2022detecting} & Swin-B & 46.9&41.7 & 45.9 & 41.7 \\
      CenterNet2 \cite{zhou2021probabilistic} &  Swin-L & 47.5 & 42.3 & 41.4 & 36.8 \\
      \midrule
      X-Paste \cite{zhao2022xpaste} & Swin-L  & 50.1 & 44.4 & 48.2 & 43.3 \\
      \multirow{2}{*}{\textbf{\method\ (Ours)}} & \multirow{2}{*}{Swin-L} & \textbf{51.2} & \textbf{45.5} & \textbf{50.1} & \textbf{45.8} \\
      & & \textcolor{red}{(+1.1)} & \textcolor{red}{(+1.1)} & \textcolor{red}{(+1.9)} & \textcolor{red}{(+2.5)} \\ 
    \bottomrule 
    \end{tabular}
  }
\caption{\textbf{Comparison with previous methods on LVIS val set.}}
\vspace{-4mm}
\label{tab:main}
\end{table}

\subsection{Ablation Studies}
We analyze the effects of the proposed strategies in \method\ through a series of ablation studies using the Swin-L~\cite{liu2021Swin} backbone.

\noindent\textbf{Effect of category diversity.}
We select 50, 250, and 566 extra categories from ImagNet-1K~\cite{russakovsky2015imagenet}, and generate 0.5k images for each category, which are added to the baseline. The baseline only uses 1,203 categories of LIVS~\cite{gupta2019lvis} to generate data. We show the results in Table~\ref{tab:category}. Generally, increasing the number of extra categories initially improves then declines model performance, peaking at 250 extra categories. The trend suggests that using extra categories to enhance category diversity can improve the model's generalization capabilities, but too many extra categories may mislead the model, leading to a decrease in performance.

\begin{table}[h]
  \small
  \centering
  \resizebox{0.9\linewidth}{!}{
      \begin{tabular}{c|cccc}
      \toprule
      \# Extra Category&$\text{AP}^{box}$&$\text{AP}^{mask}$&$\text{AP}_r^{box}$&$\text{AP}_r^{mask}$ \\
      \midrule
      0  & 49.44  & 43.75  & 42.96  & 37.91 \\
      50 & 49.92  & 44.17  & 44.94  & 39.86 \\
      250 & \textbf{50.59}  & \textbf{44.77}  & \textbf{47.99}  & \textbf{42.91} \\
      566 & 50.35  & 44.63  & 47.68  & 42.53 \\
    \bottomrule
    \end{tabular}
    }
\vspace{-2mm}
\caption{\textbf{Ablation of the number of extra categories during training.}
Using extra categories to enhance category diversity can improve the model's generalization capabilities, but too many extra categories may mislead the model, leading to a decrease in performance.}
\label{tab:category}
\end{table}

\noindent\textbf{Effect of prompt diversity.}
We select a subset of categories and use ChatGPT to generate 32 and 128 prompts for each category, with each prompt being used to generate 8 and 2 images, respectively, ensuring that the image count for each category is 0.25k. The baseline uses only one prompt per category to generate 0.25k images. The regenerated images will replace the corresponding categories in the baseline to ensure that the final data scale is consistent. The results are presented in Table~\ref{tab:prompt}. With the increase in prompt diversity, there is a continuous improvement in model performance, indicating that prompt diversity is indeed beneficial for enhancing model performance.

\begin{table}[th]
  \small
  \centering
  \resizebox{0.85\linewidth}{!}{
      \begin{tabular}{c|cccc}
      \toprule
      \# Prompt&$\text{AP}^{box}$&$\text{AP}^{mask}$&$\text{AP}_r^{box}$&$\text{AP}_r^{mask}$ \\
      \midrule
      1  & 49.65  & 44.01  & 45.68  & 41.11 \\
      32 & 50.03  & 44.39  & 45.83  & 41.32 \\
      128 & 50.27  & 44.50  & 46.49  & 41.25 \\
    \bottomrule
    \end{tabular}
    }
\vspace{-2mm}
\caption{\textbf{Ablation of the number of prompts used to generate data.}
With the increase in prompt diversity, there is a continuous improvement in model performance, indicating that prompt diversity is indeed beneficial for enhancing model performance.}
\label{tab:prompt}
\end{table}

\noindent\textbf{Effect of generative model diversity.}
We choose two commonly used generative models, Stable Diffusion~\cite{rombach2021highresolution} (SD) and DeepFloyd-IF~\cite{alex2023deepfloyd} (IF). We generate 1k images per category for each generative model, totaling 1,200k. When using a mixed dataset (SD + IF), we take 600k from SD and 600k from IF per category, respectively, to ensure the total dataset scale is consistent. The baseline does not use any generative data (none). As shown in Table~\ref{tab:model}, using data generated by either SD or IF alone can improve performance, further mixing the generative data of both leads to significant performance gains. This demonstrates that increasing model diversity is beneficial for improving model performance.

\begin{table}[h]
  \small
  \centering
  \resizebox{0.85\linewidth}{!}{
      \begin{tabular}{c|cccc}
      \toprule
      Model&$\text{AP}^{box}$&$\text{AP}^{mask}$&$\text{AP}_r^{box}$&$\text{AP}_r^{mask}$ \\
      \midrule
      none & 47.50  & 42.32  & 41.39  & 36.83 \\
      SD~\cite{rombach2021highresolution} & 48.13  & 42.82  & 43.68  & 39.15 \\
      IF~\cite{alex2023deepfloyd} & 49.44  & 43.75  & 42.96  & 37.91 \\
      SD + IF & \textbf{50.78}  & \textbf{45.27}  & \textbf{48.94}  & \textbf{44.35} \\
    \bottomrule
    \end{tabular}
    }

\caption{\textbf{Ablation of different generative models.}
Increasing model diversity is beneficial for improving model performance.}
\vspace{-2mm}
\label{tab:model}
\end{table}

\begin{figure}[h]
  \centering
  \includegraphics[width=0.96\linewidth]{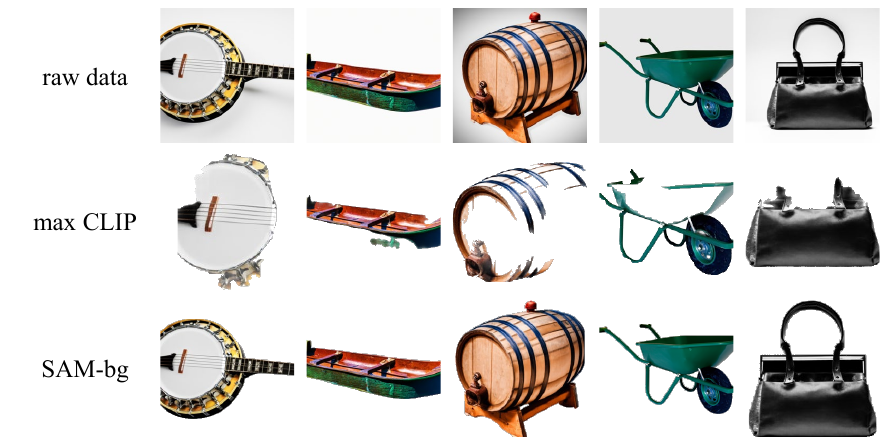}
  \caption{\textbf{Examples of object mask of different annotation strategies.}
  SAM-bg can obtain more complete and delicate masks.}
  \label{fig:mask}
\end{figure}

\noindent\textbf{Effect of annotation strategy.}
X-Paste~\cite{zhao2022xpaste} uses four models (U2Net~\cite{qin2020u2}, SelfReformer~\cite{yun2022selfreformer}, UFO~\cite{su2023unified} and CLIPseg~\cite{luddecke2022image}) to generate masks and selects the one with the highest CLIP score. We compare our proposed annotation strategy (SAM-bg) to that proposed by X-Paste (max CLIP). In Table~\ref{tab:segmentation}, SAM-bg outperforms max CLIP strategy across all metrics, indicating that our proposed strategy can produce better annotations, improving model performance. As shown in Figure~\ref{fig:mask}, SAM-bg unlocks the potential capability of SAM, obtaining precise and refined masks.

\begin{table}[h]
  \small
  \centering
  \resizebox{0.8\linewidth}{!}{
      \begin{tabular}{c|cccc}
      \toprule
      Strategy&$\text{AP}^{box}$&$\text{AP}^{mask}$&$\text{AP}_r^{box}$&$\text{AP}_r^{mask}$ \\
      \midrule
      max CLIP~\cite{zhao2022xpaste} & 49.10  & 43.45  & 42.75  & 37.55 \\
      SAM-bg & \textbf{49.44}  & \textbf{43.75}  & \textbf{42.96}  & \textbf{37.91} \\
    \bottomrule
    \end{tabular}
    }
\caption{\textbf{Ablation of different annotation strategies.}
Our proposed SAM-bg can produce better annotations, improving model performance.}
\vspace{-3mm}
\label{tab:segmentation}
\end{table}

\noindent\textbf{Effect of CLIP inter-similarity.}
We compare our proposed CLIP inter-similarity to CLIP score~\cite{zhao2022xpaste}.  The results are shown in Table~\ref{tab:similarity}. The performance of data filtered by CLIP inter-similarity is higher than that of CLIP score, demonstrating that CLIP inter-similarity can filter low-quality images more effectively.

\begin{table}[h]
  \small
  \centering
  \resizebox{0.9\linewidth}{!}{
      \begin{tabular}{c|cccc}
      \toprule
      Strategy&$\text{AP}^{box}$&$\text{AP}^{mask}$&$\text{AP}_r^{box}$&$\text{AP}_r^{mask}$ \\
      \midrule
      none & 49.44  & 43.75  & 42.96  & 37.91 \\
      CLIP score~\cite{zhao2022xpaste} & 49.84  & 44.27  & 44.83  & 40.82 \\      
      CLIP inter-similarity & \textbf{50.07}  & \textbf{44.44}  & \textbf{45.53}  & \textbf{41.16} \\
    \bottomrule
    \end{tabular}
    }
\vspace{-2mm}
\caption{\textbf{Ablation of the different filtration strategies.}
Our proposed CLIP inter-similarity can filter low-quality images more effectively.}
\vspace{-5mm}
\label{tab:similarity}
\end{table}

%% file: sec/5_conclusion.tex
\section{Conclusions}
\label{sec:conclusion}
In this paper, we explain the role of generative data augmentation from the perspective of data distribution discrepancies and find that generative data can expand the data distribution that the model can learn, mitigating overfitting the training set. Furthermore, we find that data diversity of generative data is crucial for improving model performance. Therefore, we design an efficient data diversity enhancement strategy, Generative Data Diversity Enhancement. We design various diversity enhancement strategies to increase data diversity from the aspects of category diversity, prompt diversity, and generative model diversity. Finally, we optimize the data generative pipeline by designing the annotation strategy SAM-background to obtain higher quality annotations and introducing the metric CLIP inter-similarity to filter data, which further improves the quality of the generative dataset. Through these designed strategies, our proposed method significantly outperforms the existing strong models. We hope \method\ can provide new insights and inspirations for future research on the effectiveness and efficiency of generative data augmentation.

\subsection*{Acknowledgments}
This work was in part 
support\-ed by National Key R\&D Program of China (No.\  20\-22\-ZD\-01\-18\-700).

%% file: sec/X_suppl.tex
\clearpage

\section*{Appendix}
\appendix

\section{Implementation Details}
\label{sec:implement_details}
\subsection{Data Distribution Analysis}
\label{sec:implement_details_dda}
We use the image encoder of CLIP~\cite{radford2021learning} ViT-L/14 to extract image embeddings. For objects in the LVIS~\cite{gupta2019lvis} dataset, we extract embeddings from the object regions instead of the whole images. First, we blur the regions outside the object masks using the normalized box filter, with the kernel size of (10, 10). Then, to prevent objects from being too small, we pad around the object boxes to ensure the minimum width of the padded boxes is 80 pixels, and crop the images according to the padded boxes. Finally, the cropped images are fed into the CLIP image encoder to extract embeddings. For generative images, the whole images are fed into the CLIP image encoder to extract embeddings. At last, we use UMAP~\cite{mcinnes2018umap} to reduce dimensions for visualization.
$\tau$ is set to 0.9 in the energy function.

To investigate the potential impact of noise in the rare classes to TVG metrics, we conduct additional experiments to demonstrate the validity of TVG. We randomly take five different models each for the LVIS and LVIS + Gen data sources, compute the mean ($\mu$) and standard deviation ($\sigma$) of their TVG, and calculate the 3 sigma range ($\mu+3\sigma$ and $\mu-3\sigma$), which we think represents the maximum fluctuation that potential noise could induce. As shown in Table~\ref{tab:sigma_all}, we find that: 1) The TVGs of LVIS all exceed the 3 sigma upper bound of LVIS + Gen, while the TVGs of LVIS + Gen are all below the 3 sigma lower bound of LVIS, and there is no overlap between the 3 sigma ranges of LVIS and LVIS + Gen; 2) For both LVIS + Gen and LVIS, there is no overlap between the 3 sigma ranges of different groups, e.g. frequent and common, common and rare. These two findings suggest that even in the presence of potential noise, the results can not be attributed to those fluctuations. Therefore, we think our proposed TVG metrics are reasonable and can support the conclusions.

\begin{table}[h]
  \centering
  \begin{subtable}{0.95\linewidth}
  \centering
  
  \resizebox{\linewidth}{!}{
      \begin{tabular}{c|cccccccc}
      \toprule
      ~ & $\text{TVG}_f^{box}$ & $\text{TVG}_f^{mask}$ & $\text{TVG}_c^{box}$ & $\text{TVG}_c^{mask}$ & $\text{TVG}_r^{box}$ & $\text{TVG}_r^{mask}$ \\
      \midrule
      $\mu$  & 9.98 & 8.60 & 16.59 & 13.36 & 30.23 & 24.22 \\
      $\sigma$ & 0.24 & 0.18 & 0.56 & 0.44 & 1.12 & 1.18 \\
      $\mu+3\sigma$ & 10.70 & 9.15 & 18.26 & 14.69 & 33.58 & 27.77 \\
      $\mu-3\sigma$ & 9.25 & 8.06 & 14.91 & 12.04 & 26.88 & 20.68 \\
      \midrule
      LVIS & 13.16  & 10.71  & 21.80  & 16.80  & 39.59  & 31.68 \\
      \bottomrule
    \end{tabular}
  }
  \caption{LVIS + Gen}
  \label{tab:sigma_gen}
  \end{subtable}%

  \begin{subtable}{0.95\linewidth}
  \centering
  \resizebox{\linewidth}{!}{
      \begin{tabular}{c|cccccccc}
      \toprule
      ~ & $ \text{TVG}_f^{box}$ & $\text{TVG}_f^{mask}$ & $\text{TVG}_c^{box}$ & $\text{TVG}_c^{mask}$ & $\text{TVG}_r^{box}$ & $\text{TVG}_r^{mask}$ \\
      \midrule
      $\mu$  & 13.95 & 11.40 & 22.53 & 17.16 & 43.46 & 35.10 \\
      $\sigma$ & 0.41 & 0.35 & 0.43 & 0.33 & 1.98 & 1.75 \\
      $\mu+3\sigma$ & 15.17 & 12.45 & 23.81 & 18.14 & 49.39 & 40.37 \\
      $\mu-3\sigma$ & 12.73 & 10.34 & 21.25 & 16.17 & 37.53 & 29.84 \\
      \midrule
      LVIS + Gen & 9.64  & 8.38  & 15.64  & 12.69  & 29.39  & 22.49 \\
      \bottomrule
    \end{tabular}
  }
  \caption{LVIS}
  \label{tab:sigma_lvis}
  \end{subtable}%
\caption{\textbf{Statistics of train-val gap on different data sources.}}
\label{tab:sigma_all}
\end{table}

\subsection{Category Diversity}
We compute the path similarity of WordNet~\cite{fellbaum2010wordnet} synsets between 1,000 categories in ImageNet-1K~\cite{russakovsky2015imagenet} and 1,203 categories in LVIS~\cite{gupta2019lvis}. For each of the 1,000 categories in ImageNet-1K, if the highest similarity for that category is below 0.4, we consider the category to be non-existent in LVIS and designate it as an extra category. Based on this method, 566 categories can serve as extra categories. The names of these 566 categories are presented in Table~\ref{tab:imgnet_category}.

\subsection{Prompt Diversity}
\begin{figure*}[t]
  \centering
  \includegraphics[width=\linewidth]{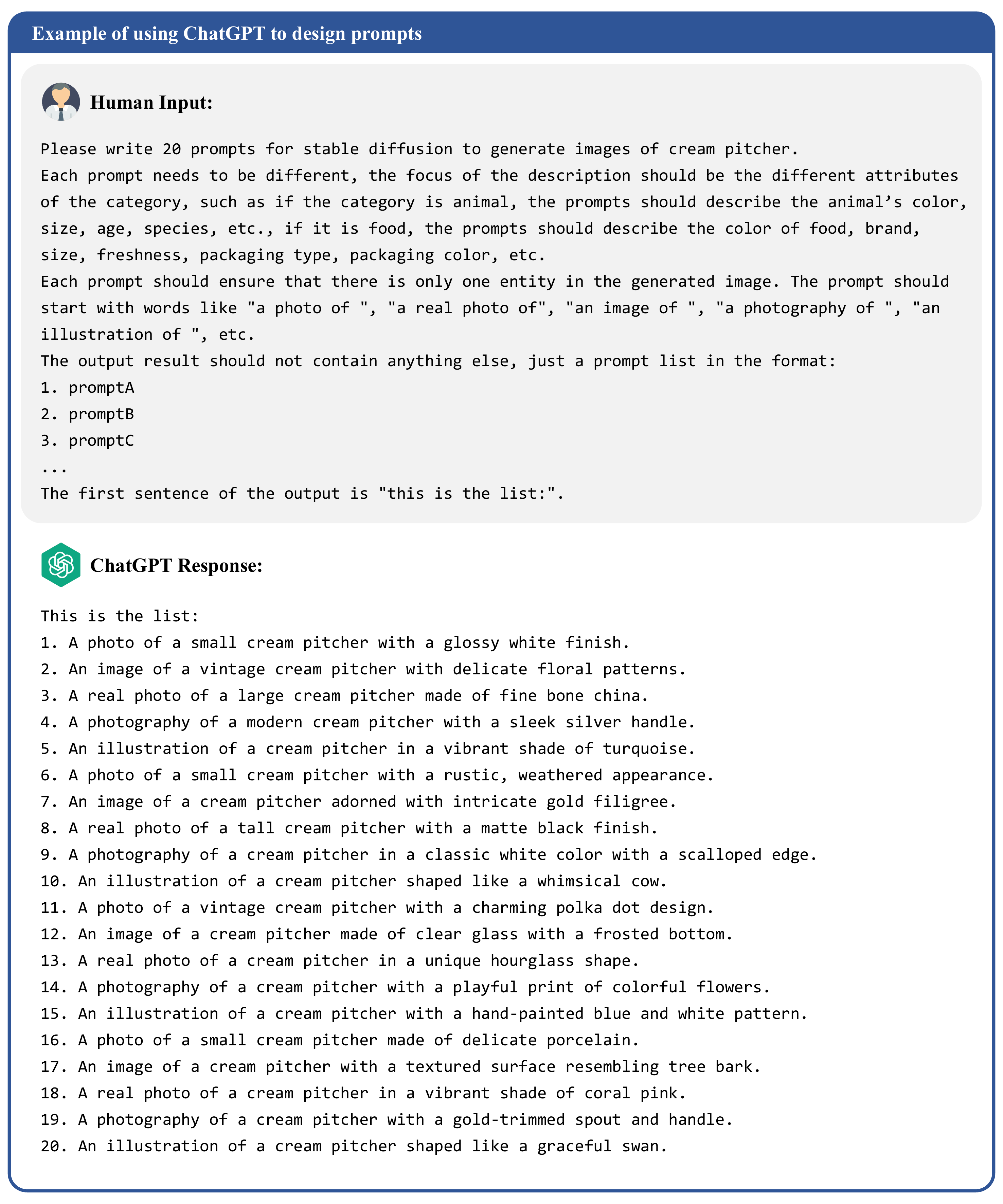}
  \caption{\textbf{Example of using ChatGPT to design prompts.}}
  \label{fig:chatgpt_prompts}
\end{figure*}

Limited by the inference cost of ChatGPT, we use the manually designed prompts as the base and only use ChatGPT to enhance the prompt diversity for a subset of categories.
For manually designed prompts, the template of prompts is ``a photo of a single \textit{\{category\_name\}}, \textit{\{category\_def\}}, in a white background". category\_name and category\_def are from LVIS~\cite{gupta2019lvis} category information.
For ChatGPT designed prompts, we select a subset of categories and use ChatGPT to enhance prompt diversity for these categories. The names of the 144 categories in this subset are shown in Table~\ref{tab:chatgpt_category}.
We use GPT-3.5-turbo and have three requirements for the ChatGPT: 1) each prompt should be as different as possible; 2) each prompt should ensure that there is only one object in the image; 3) prompts should describe different attributes of the category. Therefore, the input prompts to ChatGPT contain these three requirements. Examples of input prompts and the corresponding responses from ChatGPT are illustrated in Figure~\ref{fig:chatgpt_prompts}. To conserve output token length, there is no strict requirement for ChatGPT designed prompts to end with ``in a white background", and this constraint will be added when generating images.

\subsection{Generative Model Diversity}
We select two commonly used generative models, Stable Diffusion~\cite{rombach2021highresolution} and DeepFloyd-IF~\cite{alex2023deepfloyd}.
For Stable Diffusion, we use Stable Diffusion V1.5, with 50 inference steps and a guidance scale of 7.5. All other parameters are set to their defaults.
For DeepFloyd-IF, we use the output images from stage II, with stage I using the weight IF-I-XL-v1.0 and stage II using IF-II-L-v1.0. All parameters are set to their defaults.

\subsection{Instance Annotation}
We employ SAM~\cite{kirillov2023segment} ViT-H as the annotation model. We explore two annotation strategies, namely SAM-foreground and SAM-background.
SAM-foreground uses points sampled from foreground objects as input prompts. Specifically, we first obtain the approximate region of the foreground object based on the cross-attention map of the generative model using a threshold. Then, we use k-means++~\cite{arthur2007k} clustering to transform dense points within the foreground region into cluster centers. Next, we randomly select some points from the cluster centers as inputs to SAM. We use various metrics to evaluate the quality of the output mask and select the mask with the highest score as the final mask. However, although SAM-foreground is intuitive, it also has some limitations. Firstly, cross-attention maps of different categories require different thresholds to obtain foreground regions, making it cumbersome to choose the optimal threshold for each category. Secondly, the number of points required for SAM to output mask varies for different foreground objects. Complex object needs more points than simple object, making it challenging to control the number of points. Additionally, the position of points significantly influences the quality of SAM's output mask. If the position of points is not appropriate, this strategy is prone to generating incomplete masks.

\begin{figure}[h]
  \centering
  \includegraphics[width=0.8\linewidth]{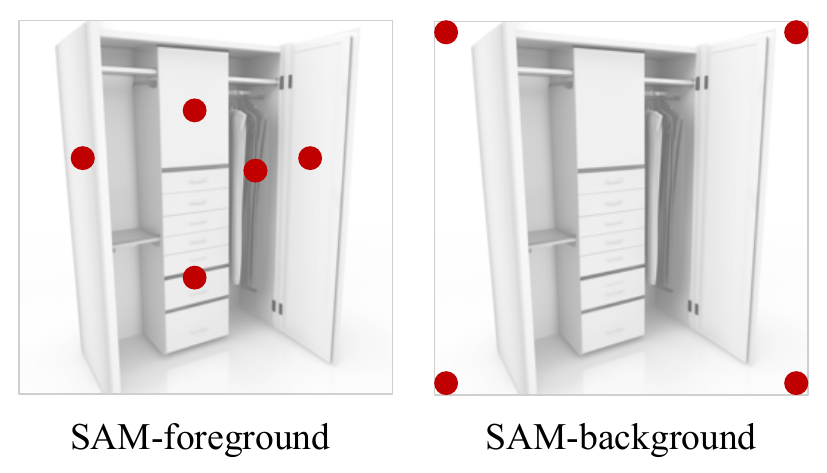}
  \caption{\textbf{Illustrations of point selection for SAM-foreground and SAM-background.}}
  \label{fig:sam_fg_bg}
\end{figure}

\begin{figure}[h]
  \centering
  \includegraphics[width=0.9\linewidth]{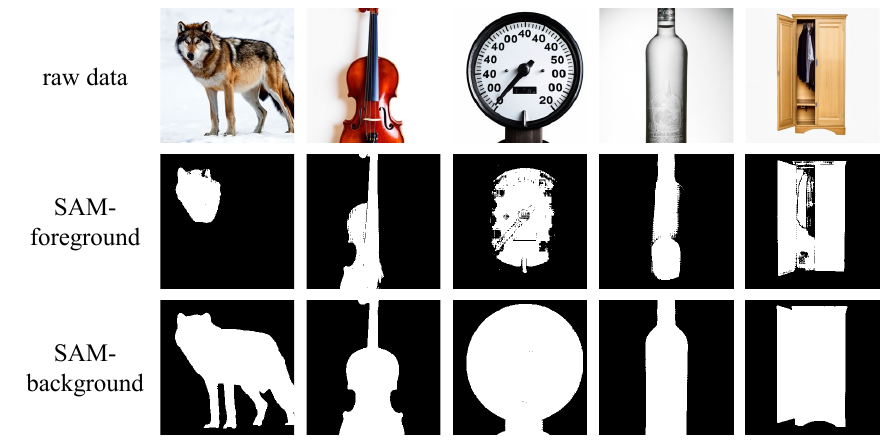}
  \caption{\textbf{Examples of annotations from SAM-foreground and SAM-background.}
  By using SAM-background for annotation, more refined masks can be obtained.}
  \label{fig:anno_fg_bg}
\end{figure}

Therefore, we discard SAM-foreground and propose a simpler and more effective annotation strategy, SAM-background. Due to our leveraging of the controllability of the generative model in instance generation, the generative images have two characteristics: 1) each image predominantly contains only one foreground object; 2) the background of the images is relatively simple. SAM-background directly uses the four corner points of the image as input prompts for SAM to obtain the background mask, then inverts the background mask as the mask of the foreground object.
The illustrations of point selection for SAM-foreground and SAM-background are shown in Figure~\ref{fig:sam_fg_bg}. By using SAM-background for annotation, more refined masks can be obtained. Examples of annotations from SAM-foreground and SAM-background are shown in Figure~\ref{fig:anno_fg_bg}. 

To further validate the effectiveness of SAM-background, we manually annotate masks for some images as ground truth (gt). We apply both strategies to annotate these images and calculate the mIoU between the resulting masks and the ground truth. The results in Table~\ref{tab:fg_bg} indicate that SAM-background achieves better annotation quality.

\begin{table}[h]
  \small
  \centering
  \resizebox{0.5\linewidth}{!}{
      \begin{tabular}{c|c}
      \toprule
      Strategy&mIoU \\
      \midrule
      SAM-foreground & 0.8163 \\
      SAM-background & \textbf{0.9418} \\
    \bottomrule
    \end{tabular}
    }
\caption{\textbf{Results of SAM-foreground and SAM-background.}
SAM-background achieves better annotation quality.}
\label{tab:fg_bg}
\end{table}

\subsection{Instance Filtration}
We use the image encoder of CLIP~\cite{radford2021learning} ViT-L/14 to extract image embeddings. The embedding extraction process is consistent with Sec~\ref{sec:implement_details_dda}. Then we calculate the cosine similarity between embeddings of objects in LVIS training set and embeddings of generative images. For each generative image, the final CLIP inter-similarity is the average similarity with all objects of the same category in the training set. Through experiments, we find that when the filtering threshold is 0.6, the model achieves the best performance and strikes a balance between data diversity and quality, so we set the threshold to 0.6.

Furthermore, we also explore other filtration strategies. From our experiments, using pure image-trained models like DINOv2~\cite{oquab2023dinov2} as image encoder or combining CLIP score and CLIP inter-similarity is not as good as using just CLIP inter-similarity alone, as shown in Table~\ref{tab:more_similarity}. Therefore, we ultimately opt to only use CLIP inter-similarity.

\begin{table}[h]
  \centering
  \resizebox{\linewidth}{!}{
      \begin{tabular}{c|cccc}
      \toprule
      Strategy&$\text{AP}^{box}$&$\text{AP}^{mask}$&$\text{AP}_r^{box}$&$\text{AP}_r^{mask}$ \\
      \midrule
      DINOv2 & 48.02  & 42.39  & 40.31  & 35.27 \\
      CLIP score + CLIP inter-similarity & 49.82  & 44.30  & 45.26  & 40.92 \\
      CLIP inter-similarity & \textbf{50.07}  & \textbf{44.44}  & \textbf{45.53}  & \textbf{41.16} \\
      \bottomrule
    \end{tabular}
  }
\caption{\textbf{Results of different filtration strategies.}}
\label{tab:more_similarity}
\end{table}

\subsection{Instance Augmentation}
In instance augmentation, we use the instance paste strategy proposed by~\citet{zhao2022xpaste} to increase model learning efficiency on generative data. Each image contains up to 20 pasted instances at most.

The parameters not specified in the paper are consistent with X-Paste~\cite{zhao2022xpaste}.

\section{Visualization}
\label{sec:visualization}
\subsection{Prompt Diversity}
\begin{figure*}[t]
  \centering
  \includegraphics[width=0.9\linewidth]{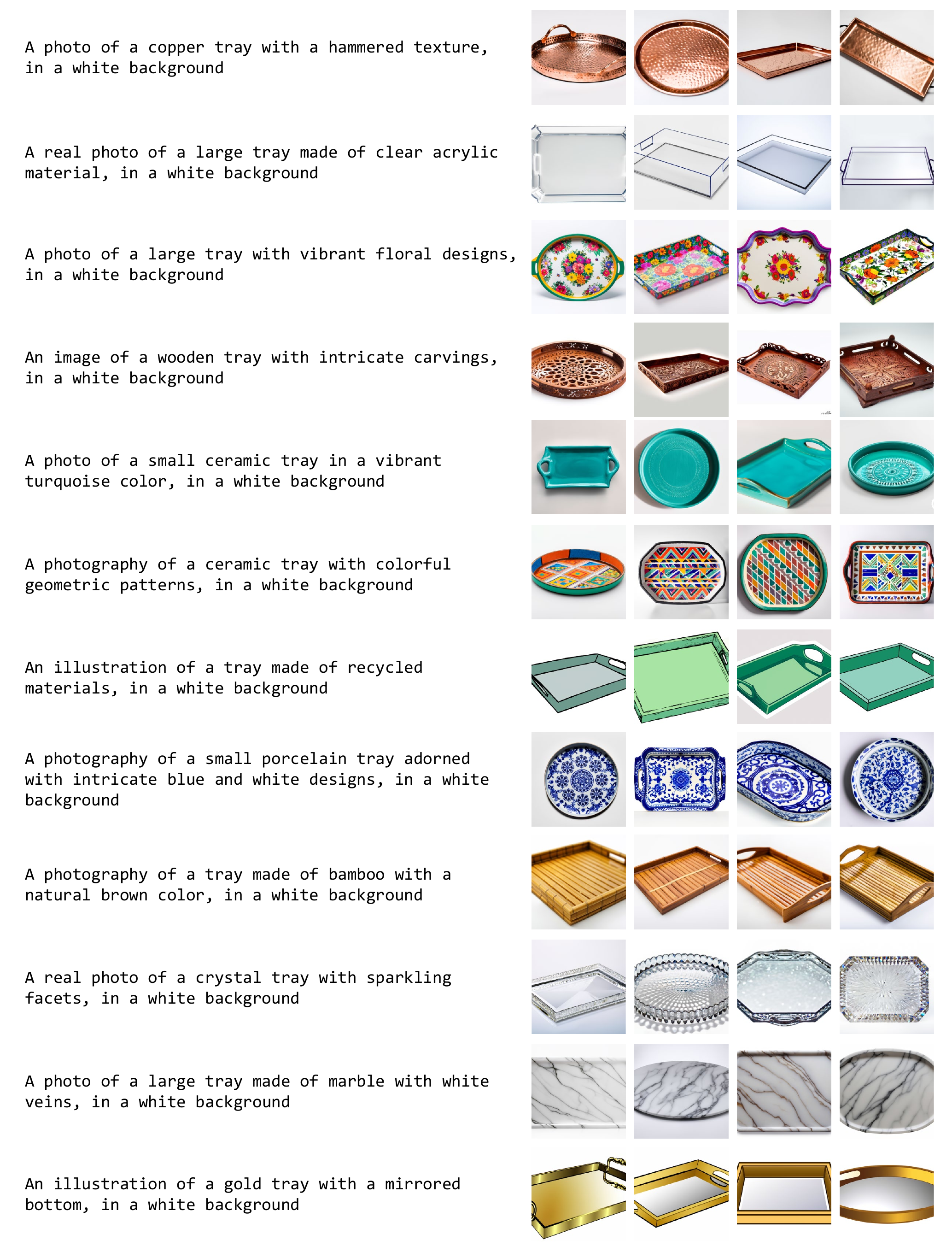}
  \caption{\textbf{Examples of ChatGPT designed prompts and corresponding generative images.}
  Images generated from ChatGPT designed prompts have diverse textures, styles, patterns, etc.}
  \label{fig:prompt_to_images}
\end{figure*}

\begin{figure*}[t]
  \centering
  \includegraphics[width=0.9\linewidth]{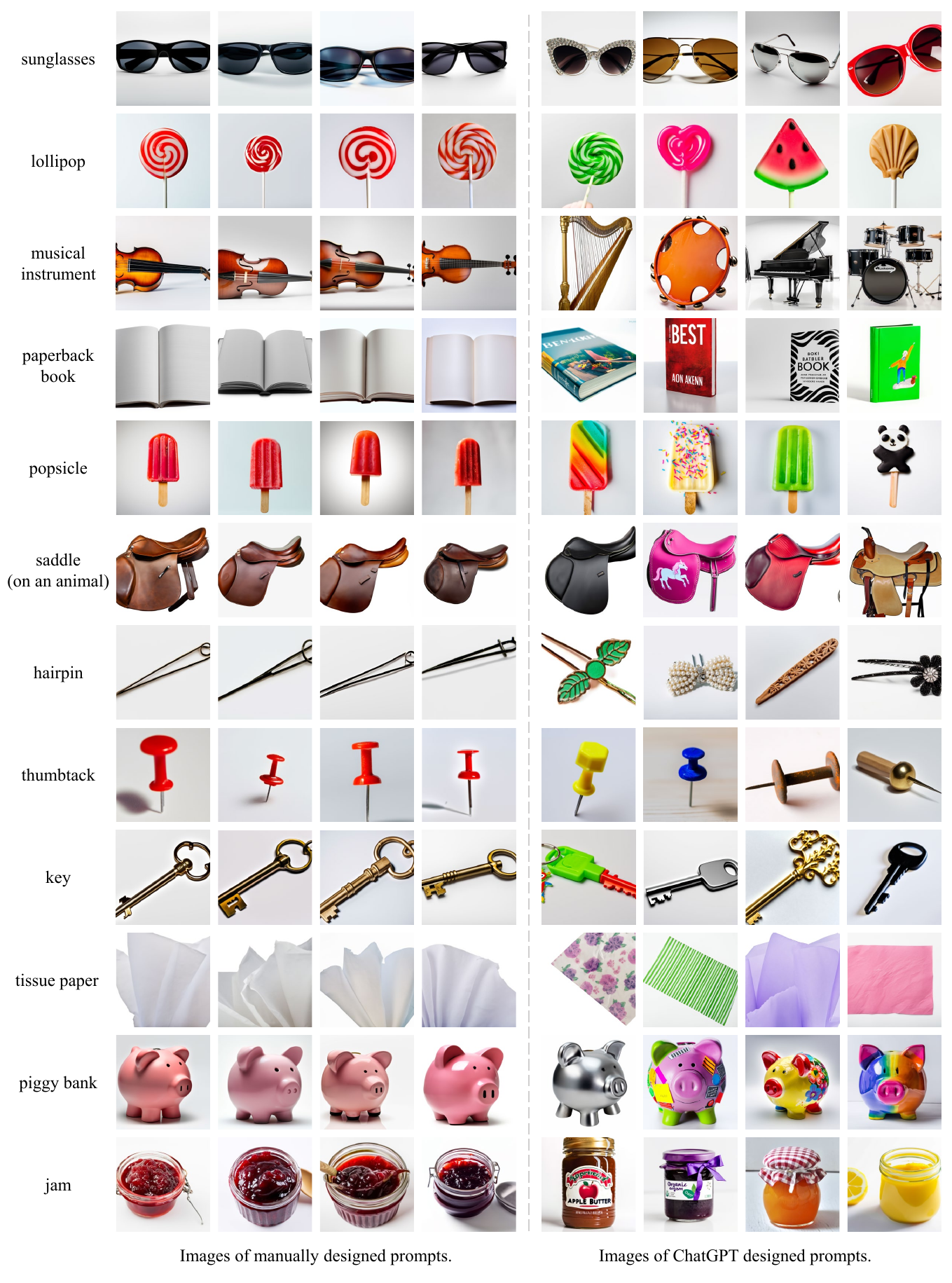}
  \caption{\textbf{Examples of generative data using different prompts.}
  By using prompts designed by ChatGPT, the diversity of generative images in terms of shapes, textures, etc. can be significantly improved.}
  \label{fig:manually_chatgpt_page}
\end{figure*}

We find that images generated from ChatGPT designed prompts have diverse textures, styles, patterns, etc., greatly enhancing data diversity. The ChatGPT designed prompts and the corresponding generative images are shown in Figure~\ref{fig:prompt_to_images}. 
Compared to manually designed prompts, the diversity of images generated from ChatGPT designed prompts can be significantly improved. A visual comparison between generative images from manually designed prompts and ChatGPT designed prompts is shown in Figure~\ref{fig:manually_chatgpt_page}.

\subsection{Generative Model Diversity}
\begin{figure*}[t]
  \centering
  \includegraphics[width=0.95\linewidth]{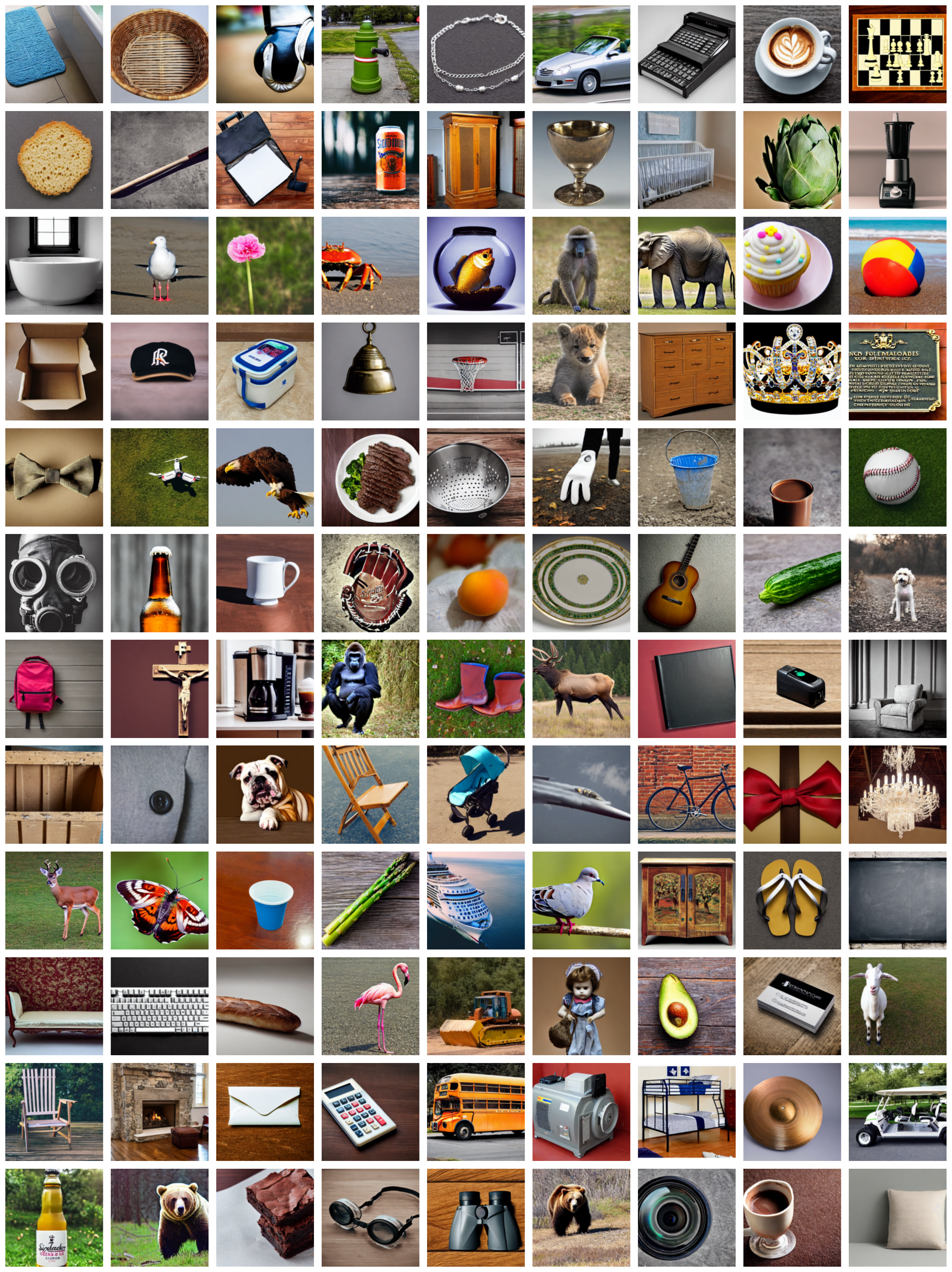}
  \caption{\textbf{Examples from Stable Diffusion.}
  The samples generated by different generative models vary, even within the same category.}
  \label{fig:sd}
\end{figure*}

\begin{figure*}[t]
  \centering
  \includegraphics[width=0.95\linewidth]{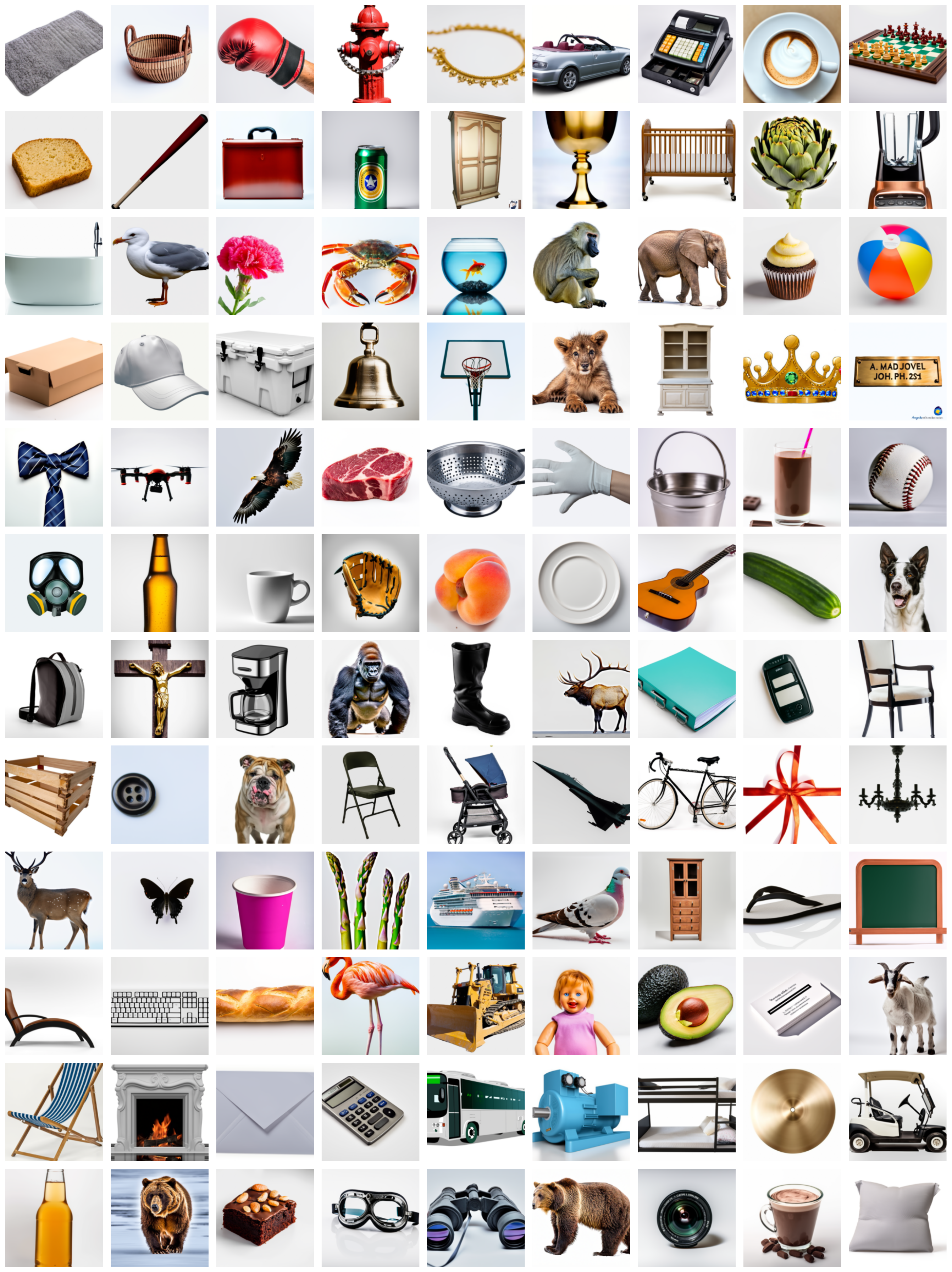}
  \caption{\textbf{Examples from DeepFloyd-IF.}
  The samples generated by different generative models vary, even within the same category.}
  \label{fig:if}
\end{figure*}

The images generated by Stable Diffusion and DeepFloyd-IF are different, even within the same category, significantly enhancing the data diversity. Both Stable Diffusion and DeepFloyd-IF are capable of producing images belonging to the target categories. However, the images generated by DeepFloyd-IF appear more photorealistic and consistent with the prompt texts. This indicates DeepFloyd-IF's superiority in image generation quality and controllability through text prompts. Examples from Stable Diffusion and DeepFloyd-IF are shown in Figure~\ref{fig:sd}  and Figure~\ref{fig:if}, respectively.

\subsection{Instance Annotation}
\begin{figure*}[t]
  \centering
  \includegraphics[width=0.9\linewidth]{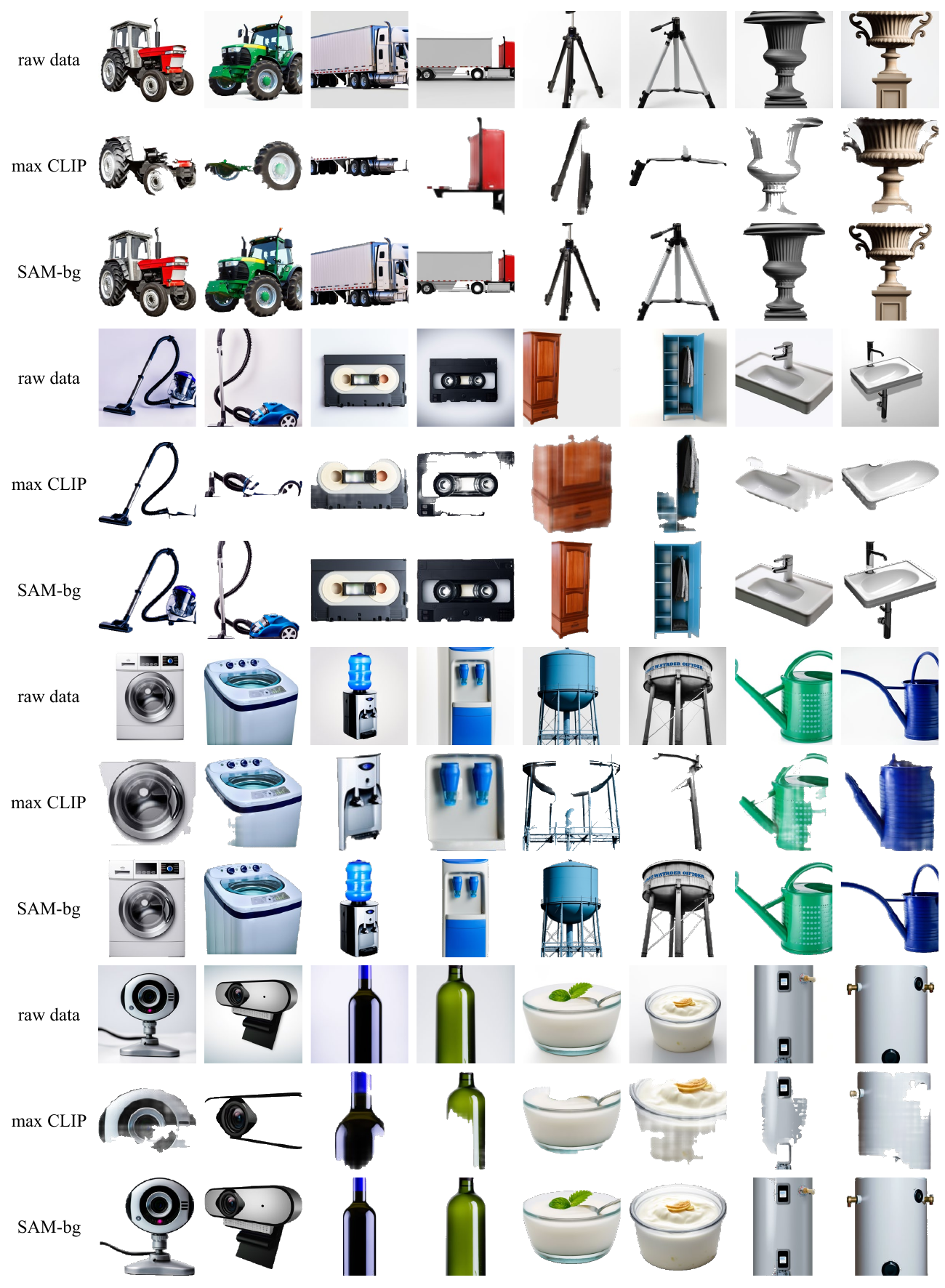}
  \caption{\textbf{Examples of different annotation strategies.}
  Masks generated by max CLIP tend to be incomplete, while our proposed SAM-bg is able to produce more refined and complete masks when processing images with multiple categories.}
  \label{fig:anno_page}
\end{figure*}

 In terms of annotation quality, masks generated by max CLIP~\cite{zhao2022xpaste} tend to be incomplete, while our proposed SAM-bg is able to produce more refined and complete masks when processing images of multiple categories. As shown in Figure~\ref{fig:anno_page}, our proposed annotation strategy can output more precise and refined masks compared to max CLIP.

\subsection{Instance Augmentation}
\begin{figure*}[t]
  \centering
  \includegraphics[width=0.9\linewidth]{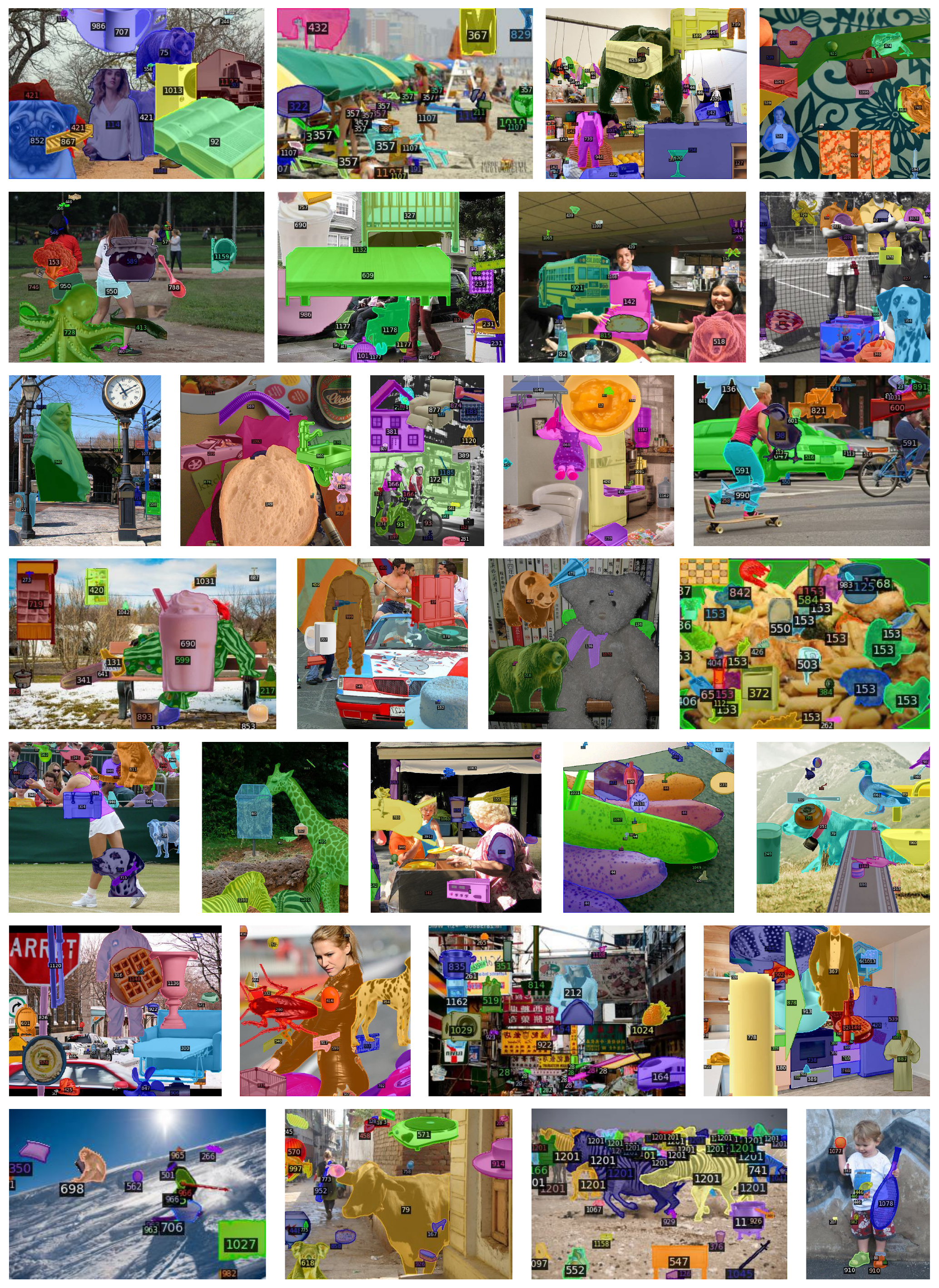}
  \caption{\textbf{Examples of augmented data.}
  The use of instance augmentation strategies helps alleviate the limitation in relatively simple scenes of generative data and improves the efficiency of model learning on the generative data.}
  \label{fig:augmentation}
\end{figure*}

The use of instance augmentation strategies helps alleviate the limitation in relatively simple scenes of generative data and improves the efficiency of model learning on the generative data. Examples of augmented data are shown in Figure~\ref{fig:augmentation}.


\small
\onecolumn
\begin{longtable}[pt]{llll}
\centering
tench & great\_white\_shark & tiger\_shark & electric\_ray\\
stingray & brambling & goldfinch & house\_finch\\
junco & indigo\_bunting & American\_robin & bulbul\\
jay & magpie & chickadee & American\_dipper\\
kite\_(bird\_of\_prey) & fire\_salamander & smooth\_newt & newt\\
spotted\_salamander & axolotl & American\_bullfrog & loggerhead\_sea\_turtle\\
leatherback\_sea\_turtle & banded\_gecko & green\_iguana & Carolina\_anole\\
desert\_grassland\_whiptail\_lizard & agama & frilled-necked\_lizard & alligator\_lizard\\
Gila\_monster & European\_green\_lizard & chameleon & Komodo\_dragon\\
Nile\_crocodile & triceratops & worm\_snake & ring-necked\_snake\\
eastern\_hog-nosed\_snake & smooth\_green\_snake & kingsnake & garter\_snake\\
water\_snake & vine\_snake & night\_snake & boa\_constrictor\\
African\_rock\_python & Indian\_cobra & green\_mamba & Saharan\_horned\_viper\\
eastern\_diamondback\_rattlesnake & sidewinder\_rattlesnake & trilobite & harvestman\\
scorpion & tick & centipede & black\_grouse\\
ptarmigan & ruffed\_grouse & prairie\_grouse & peafowl\\
quail & partridge & sulphur-crested\_cockatoo & lorikeet\\
coucal & bee\_eater & hornbill & jacamar\\
toucan & red-breasted\_merganser & black\_swan & tusker\\
echidna & platypus & wallaby & wombat\\
jellyfish & sea\_anemone & brain\_coral & flatworm\\
nematode & conch & snail & slug\\
sea\_slug & chiton & chambered\_nautilus & American\_lobster\\
crayfish & hermit\_crab & isopod & white\_stork\\
black\_stork & spoonbill & great\_egret & crane\_bird\\
limpkin & common\_gallinule & American\_coot & bustard\\
ruddy\_turnstone & dunlin & common\_redshank & dowitcher\\
oystercatcher & albatross & grey\_whale & dugong\\
sea\_lion & Chihuahua & Japanese\_Chin & Maltese\\
Pekingese & Shih\_Tzu & King\_Charles\_Spaniel & Papillon\\
toy\_terrier & Rhodesian\_Ridgeback & Afghan\_Hound & Basset\_Hound\\
Beagle & Bloodhound & Bluetick\_Coonhound & Black\_and\_Tan\_Coonhound\\
Treeing\_Walker\_Coonhound & English\_foxhound & Redbone\_Coonhound & borzoi\\
Irish\_Wolfhound & Italian\_Greyhound & Whippet & Ibizan\_Hound\\
Norwegian\_Elkhound & Otterhound & Saluki & Scottish\_Deerhound\\
Weimaraner & Staffordshire\_Bull\_Terrier & American\_Staffordshire\_Terrier & Bedlington\_Terrier\\
Border\_Terrier & Kerry\_Blue\_Terrier & Irish\_Terrier & Norfolk\_Terrier\\
Norwich\_Terrier & Yorkshire\_Terrier & Wire\_Fox\_Terrier & Lakeland\_Terrier\\
Sealyham\_Terrier & Airedale\_Terrier & Cairn\_Terrier & Australian\_Terrier\\
Dandie\_Dinmont\_Terrier & Boston\_Terrier & Miniature\_Schnauzer & Giant\_Schnauzer\\
Standard\_Schnauzer & Scottish\_Terrier & Tibetan\_Terrier & Australian\_Silky\_Terrier\\
Soft-coated\_Wheaten\_Terrier & West\_Highland\_White\_Terrier & Lhasa\_Apso & Flat-Coated\_Retriever\\
Curly-coated\_Retriever & Golden\_Retriever & Labrador\_Retriever & Chesapeake\_Bay\_Retriever\\
German\_Shorthaired\_Pointer & Vizsla & English\_Setter & Irish\_Setter\\
Gordon\_Setter & Brittany\_dog & Clumber\_Spaniel & English\_Springer\_Spaniel\\
Welsh\_Springer\_Spaniel & Cocker\_Spaniel & Sussex\_Spaniel & Irish\_Water\_Spaniel\\
Kuvasz & Schipperke & Groenendael\_dog & Malinois\\
Dobermann & Miniature\_Pinscher & Greater\_Swiss\_Mountain\_Dog & Bernese\_Mountain\_Dog\\
Appenzeller\_Sennenhund & Entlebucher\_Sennenhund & Boxer & Bullmastiff\\
Tibetan\_Mastiff & Great\_Dane & St.\_Bernard & husky\\
Alaskan\_Malamute & Siberian\_Husky & Affenpinscher & Samoyed\\
Pomeranian & Chow\_Chow & Keeshond & brussels\_griffon\\
Pembroke\_Welsh\_Corgi & Cardigan\_Welsh\_Corgi & Toy\_Poodle & Miniature\_Poodle\\
Standard\_Poodle & dingo & dhole & African\_wild\_dog\\
hyena & red\_fox & kit\_fox & Arctic\_fox\\
grey\_fox & tabby\_cat & tiger\_cat & Persian\_cat\\
Siamese\_cat & Egyptian\_Mau & lynx & leopard\\
snow\_leopard & jaguar & cheetah & mongoose\\
meerkat & dung\_beetle & rhinoceros\_beetle & fly\\
bee & ant & grasshopper & cricket\_insect\\
stick\_insect & praying\_mantis & cicada & leafhopper\\
lacewing & damselfly & red\_admiral\_butterfly & monarch\_butterfly\\
small\_white\_butterfly & sea\_urchin & sea\_cucumber & hare\\
fox\_squirrel & guinea\_pig & wild\_boar & warthog\\
ox & water\_buffalo & bison & bighorn\_sheep\\
Alpine\_ibex & hartebeest & impala\_(antelope) & llama\\
weasel & mink & black-footed\_ferret & otter\\
skunk & badger & armadillo & three-toed\_sloth\\
orangutan & chimpanzee & gibbon & siamang\\
guenon & patas\_monkey & macaque & langur\\
black-and-white\_colobus & proboscis\_monkey & marmoset & white-headed\_capuchin\\
howler\_monkey & titi\_monkey & Geoffroy's\_spider\_monkey & common\_squirrel\_monkey\\
ring-tailed\_lemur & indri & red\_panda & snoek\_fish\\
eel & rock\_beauty\_fish & clownfish & sturgeon\\
gar\_fish & lionfish & academic\_gown & accordion\\
aircraft\_carrier & altar & apiary & assault\_rifle\\
bakery & balance\_beam & baluster\_or\_handrail & barbershop\\
barn & barometer & bassinet & bassoon\\
lighthouse & bell\_tower & baby\_bib & boathouse\\
bookstore & breakwater & breastplate & butcher\_shop\\
carousel & tool\_kit & automated\_teller\_machine & cassette\_player\\
castle & catamaran & cello & chain\\
chain-link\_fence & chainsaw & chiffonier & Christmas\_stocking\\
church & movie\_theater & cliff\_dwelling & cloak\\
clogs & spiral\_or\_coil & candy\_store & cradle\\
construction\_crane & croquet\_ball & cuirass & dam\\
desktop\_computer & disc\_brake & dock & dome\\
drilling\_rig & electric\_locomotive & entertainment\_center & face\_powder\\
fire\_screen & flute & fountain & French\_horn\\
gas\_pump & golf\_ball & gong & greenhouse\\
radiator\_grille & grocery\_store & guillotine & hair\_spray\\
half-track & hand-held\_computer & hard\_disk\_drive & harmonica\\
harp & combine\_harvester & holster & home\_theater\\
honeycomb & hook & gymnastic\_horizontal\_bar & jigsaw\_puzzle\\
knot & lens\_cap & library & lifeboat\\
lighter & lipstick & lotion & loupe\_magnifying\_glass\\
sawmill & messenger\_bag & maraca & marimba\\
mask & matchstick & maypole & maze\\
megalith & military\_uniform & missile & mobile\_home\\
modem & monastery & monitor & moped\\
mortar\_and\_pestle & mosque & mosquito\_net & tent\\
mousetrap & moving\_van & muzzle & metal\_nail\\
neck\_brace & notebook\_computer & obelisk & oboe\\
ocarina & odometer & oil\_filter & pipe\_organ\\
oscilloscope & oxygen\_mask & palace & pan\_flute\\
parallel\_bars & patio & pedestal & photocopier\\
plectrum & Pickelhaube & picket\_fence & pier\\
pirate\_ship & block\_plane & planetarium & plastic\_bag\\
plate\_rack & plunger & police\_van & prayer\_rug\\
prison & hockey\_puck & punching\_bag & purse\\
radio & radio\_telescope & rain\_barrel & fishing\_casting\_reel\\
restaurant & rugby\_ball & safe & scabbard\\
schooner & CRT\_monitor & seat\_belt & shoe\_store\\
shoji\_screen\_or\_room\_divider & balaclava\_ski\_mask & slide\_rule & sliding\_door\\
slot\_machine & snorkel & keyboard\_space\_bar & spatula\\
motorboat & spider\_web & spindle & stage\\
steam\_locomotive & through\_arch\_bridge & steel\_drum & stethoscope\\
stone\_wall & tram & stretcher & stupa\\
submarine & sundial & sunglasses & sunscreen\\
suspension\_bridge & swing & tape\_player & television\\
thatched\_roof & threshing\_machine & throne & tile\_roof\\
tobacco\_shop & toilet\_seat & torch & totem\_pole\\
toy\_store & trimaran & triumphal\_arch & trombone\\
turnstile & typewriter\_keyboard & vaulted\_or\_arched\_ceiling & velvet\_fabric\\
vestment & viaduct & sink & whiskey\_jug\\
whistle & window\_screen & window\_shade & airplane\_wing\\
wool & split-rail\_fence & shipwreck & sailboat\\
yurt & website & crossword & dust\_jacket\\
menu & plate & guacamole & trifle\\
baguette & cabbage & broccoli & spaghetti\_squash\\
acorn\_squash & butternut\_squash & cardoon & mushroom\\
Granny\_Smith\_apple & jackfruit & cherimoya\_(custard\_apple) & pomegranate\\
hay & carbonara & chocolate\_syrup & dough\\
meatloaf & pot\_pie & red\_wine & espresso\\
tea\_cup & eggnog & mountain & bubble\\
cliff & coral\_reef & geyser & lakeshore\\
promontory & sandbar & beach & valley\\
volcano & baseball\_player & bridegroom & scuba\_diver\\
rapeseed & daisy & yellow\_lady's\_slipper & corn\\
acorn & rose\_hip & horse\_chestnut\_seed & coral\_fungus\\
gyromitra & stinkhorn\_mushroom & earth\_star\_fungus & hen\_of\_the\_woods\_mushroom\\
bolete & corn\_cob\\
\caption{\textbf{Extra categories from ImageNet-1K.}}
\label{tab:imgnet_category}\\
\end{longtable}
\normalsize
\twocolumn

\small
\onecolumn
\begin{longtable}[t]{llll}
\centering
Bible & pirate\_flag & bookmark & bow\_(weapon)\\
bubble\_gum & elevator\_car & chocolate\_mousse & compass\\
corkboard & cougar & cream\_pitcher & cylinder\\
dollar & dolphin & eyepatch & fruit\_juice\\
golf\_club & handcuff & hockey\_stick & popsicle\\
pan\_(metal\_container) & pew\_(church\_bench) & piggy\_bank & pistol\\
road\_map & satchel & sawhorse & shawl\\
sparkler\_(fireworks) & spider & string\_cheese & Tabasco\_sauce\\
turtleneck\_(clothing) & violin & waffle\_iron & whistle\\
wind\_chime & headstall\_(for\_horses) & fishing\_rod & coat\_hanger\\
clasp & crab\_(animal) & flamingo & stirrup\\
machine\_gun & pin\_(non\_jewelry) & spear & drumstick\\
cornet & bottle\_opener & easel & dumbbell\\
garden\_hose & money & saddle\_(on\_an\_animal) & garbage\\
windshield\_wiper & needle & liquor & bamboo\\
armor & pretzel & tongs & ski\_pole\\
frog & hairpin & tripod & flagpole\\
hose & belt\_buckle & streetlight & coleslaw\\
antenna & hook & Lego & thumbtack\\
coatrack & plow\_(farm\_equipment) & vinegar & strap\\
poker\_(fire\_stirring\_tool) & cufflink & chopstick & salad\\
dragonfly & musical\_instrument & sharpener & bat\_(animal)\\
lanyard & mat\_(gym\_equipment) & gargoyle & underdrawers\\
paperback\_book & razorblade & earring & sword\\
shovel & turkey\_(food) & ambulance & pencil\\
weathervane & trampoline & applesauce & jam\\
ski & tray & tissue\_paper & lamppost\\
clipboard & router\_(computer\_equipment) & battery & lollipop\\
crayon & latch & fig\_(fruit) & sunglasses\\
toothpick & business\_card & padlock & asparagus\\
shot\_glass & sled & key & bolt\\
pipe & steering\_wheel & deck\_chair & green\_bean\\
pouch & telephone\_pole & fire\_hose & ladle\\
pliers & hair\_curler & handle & screwdriver\\
dining\_table & cart & oar & wolf\\
envelope & legume & shopping\_cart & trench\_coat\\
\caption{\textbf{Categories of ChatGPT designed prompts.}}
\label{tab:chatgpt_category}\\
\end{longtable}
\normalsize
\twocolumn

